\documentclass[11pt,letterpaper]{article}
\usepackage[margin=1in]{geometry}
\usepackage{times}
\usepackage{natbib}
\setcitestyle{authoryear,round,citesep={;},aysep={,},yysep={;}}
\usepackage[T1]{fontenc}
\usepackage{amsmath,amssymb,amsthm,mathtools,bm}
\usepackage{aliascnt}
\usepackage{enumitem}
\usepackage{graphicx,booktabs}
\usepackage{microtype}
\usepackage{algorithm}
\usepackage[noend]{algpseudocode}
\usepackage[hidelinks]{hyperref}
\hypersetup{pdftitle={Bootstrap Inference for Riemannian Optimization Problems},pdfauthor={Xiaotian Chang, Qirui Hu, Yangdi Jiang, Nan Qiao}}
\usepackage[nameinlink,capitalise,noabbrev]{cleveref}
\allowdisplaybreaks
\usepackage{xcolor}
\DeclareRobustCommand{\tfill}[1]{\begingroup#1\endgroup}
\newtheorem{theorem}{Theorem}[section]

\newaliascnt{proposition}{theorem}
\newtheorem{proposition}[proposition]{Proposition}
\aliascntresetthe{proposition}

\newaliascnt{lemma}{theorem}
\newtheorem{lemma}[lemma]{Lemma}
\aliascntresetthe{lemma}

\newaliascnt{corollary}{theorem}
\newtheorem{corollary}[corollary]{Corollary}
\aliascntresetthe{corollary}

\theoremstyle{definition}
\newaliascnt{assumption}{theorem}
\newtheorem{assumption}[assumption]{Assumption}
\aliascntresetthe{assumption}

\newaliascnt{definition}{theorem}
\newtheorem{definition}[definition]{Definition}
\aliascntresetthe{definition}

\crefname{theorem}{Theorem}{Theorems}
\Crefname{theorem}{Theorem}{Theorems}
\crefname{proposition}{Proposition}{Propositions}
\Crefname{proposition}{Proposition}{Propositions}
\crefname{lemma}{Lemma}{Lemmas}
\Crefname{lemma}{Lemma}{Lemmas}
\crefname{corollary}{Corollary}{Corollaries}
\Crefname{corollary}{Corollary}{Corollaries}
\crefname{assumption}{Assumption}{Assumptions}
\Crefname{assumption}{Assumption}{Assumptions}
\crefname{definition}{Definition}{Definitions}
\Crefname{definition}{Definition}{Definitions}

\floatstyle{ruled}
\restylefloat{algorithm}
\algrenewcommand\algorithmicrequire{\textbf{Require:}}
\crefname{algorithm}{Algorithm}{Algorithms}
\Crefname{algorithm}{Algorithm}{Algorithms}

\newcommand{\M}{\mathcal M}
\newcommand{\R}{\mathbb R}
\newcommand{\Exp}{\mathbb E}
\newcommand{\Prob}{\mathbb P}
\newcommand{\F}{\mathcal F}
\newcommand{\G}{\mathcal G}
\newcommand{\ot}{\otimes}
\newcommand{\cL}{\mathcal L}
\newcommand{\grad}{\operatorname{grad}}
\newcommand{\Hess}{\operatorname{Hess}}
\newcommand{\Log}{\operatorname{Log}}
\newcommand{\Expmap}{\operatorname{Exp}}
\newcommand{\tr}{\operatorname{tr}}
\newcommand{\vecop}{\operatorname{vec}}
\newcommand{\norm}[1]{\left\lVert #1\right\rVert}
\newcommand{\ip}[2]{\left\langle #1,#2\right\rangle}
\newcommand{\ind}{\mathbf 1}
\newcommand{\BL}{\mathrm{BL}}
\newcommand{\dagg}{\dagger}
\newcommand{\eps}{\varepsilon}
\newcommand{\Cov}{\operatorname{Cov}}
\newcommand{\Gr}{\operatorname{Gr}}
\newcommand{\SPD}{\mathcal P}
\newcommand{\trans}{\mathsf T}
\title{
Bootstrap Inference for
Riemannian Optimization Problems} 
\author{%
Xiaotian Chang\thanks{Division of Mathematical Sciences, Nanyang Technological University.}\quad
Qirui Hu\thanks{School of Statistics and Data Science, Shanghai University of Finance and Economics. Corresponding author. Email: \texttt{huqirui@mail.shufe.edu.cn}.}\quad
Yangdi Jiang\footnotemark[1]\quad
Nan Qiao\thanks{School of Mathematics, Statistics and Mechanics, Beijing University of Technology, China.}%
}
\date{}

\begin{document}
\maketitle
\begin{abstract}{
We develop an online multiplier bootstrap for manifold-valued parameters estimated by averaging the iterates of Riemannian stochastic approximation. The method runs weighted iterations alongside the original algorithm, with each path using the same observations while updating its own state and intrinsic average. The weighted averages provide uncertainty assessments without a plug-in asymptotic covariance estimator. Under local stability and regularity conditions, we establish bootstrap consistency and show that a fixed number of paths suffices for asymptotically valid Student intervals and intrinsic confidence regions. The theory accommodates samplewise nonsmooth scores and identifies a critical oracle-bias regime in which the estimator’s limiting distribution shifts but the centered bootstrap distribution does not. Applications include spherical means, positive-definite matrix means and medians, and principal-subspace estimation. Simulations show coverage close to nominal in regular settings, and an analysis of North Atlantic storm locations illustrates descriptive online inference for a spherical center.

}\end{abstract}

\section{Introduction}\label{sec:intro}
Many stochastic optimization problems have parameters that are not naturally represented by unconstrained vectors.  Principal subspaces lie on Grassmann manifolds, covariance and diffusion tensors lie on the cone of positive-definite matrices, and directional parameters lie on spheres.  Retraction-based stochastic approximation respects these constraints without sacrificing sequential computation \citep{AbsilEtAl2008,Bonnabel2013}; iterative intrinsic averaging is the corresponding Polyak-Ruppert construction and can recover root-$n$ statistical accuracy under local stability and smoothness \citep{Ruppert1988,PolyakJuditsky1992,TripuraneniEtAl2018}.

The inferential obstacle is estimating the asymptotic covariance. For a regular root-$n$ averaged estimator, the limiting covariance has the form 
\[  
V=H^{-1}Q(H^{-1})^{\dagg},
\]
where $H$ is the derivative of the population vector field, $Q$ is the covariance operator of the noise term at the target, and $A^{\dagg}$ denotes the adjoint of a linear operator $A$.  A plug-in estimator must estimate both objects in compatible coordinates, which is difficult when the stochastic field is samplewise nonsmooth, when $H$ contains nuisance structure, or when a streaming procedure cannot retain matrix summaries.  Geometry adds a second obstacle: fields and derivatives at different iterates belong to different tangent spaces.

We construct an online bootstrap by running randomly weighted copies of the original algorithm alongside it. Each copy has independent weights and its own estimate and intrinsic average, but shares the observations and auxiliary random inputs with the original algorithm; variation among the weighted averages provides the uncertainty assessment. Since two manifold estimates cannot generally be subtracted as vectors, we compare them in the tangent space at the observed average and prove that this computable comparison reproduces the estimator's sampling fluctuation to first order.

We make four contributions.
\begin{enumerate}[label=(\roman*),leftmargin=2.2em]
\item We establish valid online bootstrap inference for averaged Riemannian stochastic approximation without estimating an asymptotic covariance; under local stability and regularity conditions, the theory accommodates samplewise nonsmooth scores and a nonsymmetric population derivative.
\item We justify comparing weighted averages at the observed estimate rather than at the unknown target. This resolves the need to compare different tangent spaces and makes the bootstrap statistic computable.  
\item We show that a fixed number of weighted copies suffices for asymptotically valid Student intervals and intrinsic confidence regions. As the number of copies grows, the same construction yields consistent conditional quantiles and, under additional integrability, covariance estimates.
\item We specialize the theory to spherical means, positive-definite matrix means and medians, and principal subspaces. For approximate updates, we identify when accumulated bias is negligible and when it shifts the estimator's limit without being reproduced by the centered bootstrap.
\end{enumerate}
For Euclidean smooth averaged SGD with exponential weights, our iteration and fixed-$B$ scalar interval reduce exactly to the Cheap Online Bootstrap (COnB) of \citet{LamWang2026}; our results extend this construction to averaged estimates on manifolds of fixed dimension through a local theory.

The numerical evidence follows this local scope: regular designs on three geometries have coverage near the nominal level, the localized skew spherical experiment gives coverage closer to nominal than a matched offline trajectory bootstrap at $n=100$, and the IBTrACS analysis shows the same paths providing descriptive online uncertainty for a spherical location summary.

\subsection{Related work}\label{sec:related-work}

Full-path multipliers avoid explicit sandwich estimation: \citet{FangXuYang2018} evolve perturbed stochastic-gradient paths beside the original iteration, evaluating each score at its own perturbed state, and \citet{LamWang2026} combine this construction with a fixed number of paths and a Student pivot, so that COnB is valid for smooth strongly convex averaged SGD even when $B$ is fixed.  Other Euclidean approaches estimate covariance online \citep{ZhuChenWu2023}, use batch means for nonsmooth or nonmonotone stochastic approximation \citep{JiangEtAl2025}, treat local nonconvex inference \citep{ZhongKuffnerLahiri2023}, or establish quantitative approximations for linear stochastic approximation and smooth SGD \citep{SamsonovEtAl2024,SheshukovaEtAl2026}.  In the derivative-free setting, \citet{JinXiaoBalasubramanian2021} derive a biased central limit theorem for Euclidean zeroth-order Polyak-Ruppert averages and estimate its covariance online.  Our approximation-bias analysis instead concerns manifold full-path multipliers and the cumulative predictable bias relative to the unsmoothed target.  Offline bootstrap theory for manifold-valued means is classical \citep{BhattacharyaPatrangenaru2005}, but it does not resolve comparison across tangent spaces for a continuously updated stochastic-approximation path.

Streaming principal-component inference is the closest geometric comparison: existing results give high-dimensional approximations and online multiplier procedures for the rank-one Oja error \citep{LundeSarkarWard2021}, entrywise inference with subsampling alternatives \citep{KumarPandeySarkar2025}, and Gaussian and multiplier approximations for the final general-rank Oja iterate \citep{XuLi2026}.  These results exploit PCA structure and target final iterates in high-dimensional regimes, whereas we study averaged iterations on a general manifold of fixed dimension, with Grassmann PCA as one specialization.

\begingroup Riemannian stochastic approximation and iterative averaging provide the geometric foundation \citep{Bonnabel2013,TripuraneniEtAl2018}; the additional task here is to compare complete weighted averages at the estimated base point with an influence representation in a fixed tangent space, which requires separate control of state stability, full-average localization, and geometric remainders. For geometric medians, \citet{ChengXiongLin2024} already develop online bootstrap inference using averaged stochastic gradients. Our median application treats the Riemannian normalized-distance score under explicit population-field and localization conditions.\endgroup

The rest of the paper is organized as follows. \Cref{sec:method} defines the online paths and intervals, \Cref{sec:theory} establishes their guarantees, and \Cref{sec:simulations} evaluates finite-sample coverage; the appendices develop \hyperref[sec:primitive]{primitive conditions}, \hyperref[sec:largeB-theory]{large-replication inference}, and \hyperref[sec:applications]{geometric applications}, alongside proofs, simulation details, and the North Atlantic storm analysis.

\section{Methodology}\label{sec:method}

\subsection{Riemannian stochastic approximation and intrinsic averaging}\label{sec:base-method}

Let $(\M,g)$ be a connected $d$-dimensional Riemannian manifold, and let $x_\star\in\M$ be a locally unique zero of a population vector field.  Write $d(x,y)$ for geodesic distance ($d$ alone is the manifold dimension), $T_x\M$ for the tangent space at $x$ with Riemannian norm $\norm{\cdot}_x$, $R_x$ for a retraction with locally defined inverse $R_x^{-1}$, and $\mathcal P_{x\to y}$ for parallel transport along the unique minimizing geodesic when $x,y$ lie in a common normal neighborhood.

Let $\Psi_k(x;D_k)\in T_x\M$ be the stochastic oracle, the stochastic vector-field estimate used to update $x$; its input $D_k=(Z_k,S_k)$ consists of the observation $Z_k$ and an auxiliary random input $S_k$ generated at step $k$, such as a sampled direction, minibatch, or solver seed.
The original iteration is
\begin{equation}\label{eq:base-update}
  x_k=R_{x_{k-1}}\{-\eta_k\Psi_k(x_{k-1};D_k)\},
  \qquad \eta_k=\eta_0 k^{-\alpha},\quad \eta_0>0,\quad \alpha\in(1/2,1).
\end{equation}
Its iterative intrinsic average is
\begin{equation}\label{eq:base-average}
  \bar x_1=x_1,
  \qquad
  \bar x_k=R_{\bar x_{k-1}}\left\{\frac1kR_{\bar x_{k-1}}^{-1}(x_k)\right\},
  \qquad k\ge2.
\end{equation}
In Euclidean space, \eqref{eq:base-average} computes the running arithmetic mean $\bar x_k=k^{-1}\sum_{j=1}^k x_j$.

\subsection{Full-path multiplier trajectories}\label{sec:weighted-method}

Let $(W_k)_{k\ge1}$ be i.i.d., independent of the complete data and oracle-randomness stream, with
\begin{equation}\label{eq:weight-normalization}
  \Exp W_k=1,
  \qquad \Exp(W_k-1)^2=1.
\end{equation}
A multiplier path starts from $x_0^\star=x_0$ and evolves according to
\begin{align}
  x_k^\star
  &=R_{x_{k-1}^\star}\{-\eta_k W_k\Psi_k(x_{k-1}^\star;D_k)\},
  \label{eq:weighted-update}\\
  \bar x_1^\star&=x_1^\star,
  \qquad
  \bar x_k^\star
  =R_{\bar x_{k-1}^\star}\left\{\frac1kR_{\bar x_{k-1}^\star}^{-1}(x_k^\star)\right\}.
  \label{eq:weighted-average}
\end{align}
The original and multiplier paths share $D_k$, but each path evaluates the oracle at its own state $x_{k-1}^\star$; this distinguishes the full-path procedure from a linearized resampling scheme that only reweights stored gradients.
\subsection{Intervals and implementation}\label{sec:calibration-main}
For a differentiable scalar functional $\phi$ and fixed $B\ge1$, run $B$ multiplier paths with independent weights. With confidence error probability $\tau\in(0,1)$, define
\begin{align}
  s_{\phi,n,B}^2
  &=\frac1B\sum_{b=1}^B
  \{\phi(\bar x_n^{\star(b)})-\phi(\bar x_n)\}^2,\label{eq:fixedB-scale}\\
  I_{\phi,n,B}(1-\tau)
  &=\left[\phi(\bar x_n)-t_{B,1-\tau/2}s_{\phi,n,B},
          \ \phi(\bar x_n)+t_{B,1-\tau/2}s_{\phi,n,B}\right].\label{eq:fixedB-CI}
\end{align}
Here $t_{B,1-\tau/2}$ is the Student quantile with $B$ degrees of freedom; the differences are centered at the original estimate. If $s_{\phi,n,B}=0$, report $\R$. For intrinsic regions, use the second moment of the vectors $\sqrt nR_{\bar x_n}^{-1}(\bar x_n^{\star(b)})$ and the Hotelling-type critical value $dB F_{d,B-d+1;1-\tau}/(B-d+1)$, with $B\ge d$, where $F_{\nu_1,\nu_2;q}$ is the $q$-quantile of the $F$ distribution. The completion is defined below; the singularity fallback and the intrinsic and large-$B$ regions are specified in \Cref{sec:outputs}. When $B=B_n\to\infty$, empirical conditional quantiles yield the basic interval \eqref{eq:basic-CI}.

All paths share each observation and oracle seed; any safeguard is applied after the complete oracle is multiplied, and each path updates its own full average with the required inverse maps defined along its history. Paths for a chosen horizon are initialized before the stream begins. \Cref{alg:fullpath} gives the complete procedure with reporting fallbacks; for an $O(d)$ state representation, the extra storage is $O(Bd)$ and each observation costs $B$ additional oracle evaluations and retractions.

\begin{algorithm}[!htb]
\caption{Online full-path multiplier inference}\label{alg:fullpath}
\begin{algorithmic}[1]
\Require Initial state $x_{0}$; steps $(\eta_k)$; retraction and local inverse;
\Statex oracle and any state safeguard; $B$ paths; multiplier law; reporting rule.
\State Initialize $x_0^{\star(b)}\gets x_0$ for $b=1,\ldots,B$.
\For{$k=1,\ldots,n$}
  \State Read $Z_k$, generate the oracle seed $S_k$, and set $D_k=(Z_k,S_k)$.
  \State Evaluate $\Psi_k(x_{k-1};D_k)$ and update $x_k$ by \eqref{eq:base-update}, including any specified safeguard.
    \State If $k=1$, set $\bar x_1\gets x_1$; otherwise update $\bar x_k$ by \eqref{eq:base-average}.
  \For{$b=1,\ldots,B$}
    \State Draw $W_k^{(b)}$ independently of the original stream and other weights.
    \State Evaluate $\Psi_k(x_{k-1}^{\star(b)};D_k)$ at this path's own state.
    \State Update $x_k^{\star(b)}$ by \eqref{eq:weighted-update}, multiplying the entire oracle before applying any specified safeguard.
    \State If $k=1$, set $\bar x_1^{\star(b)}\gets x_1^{\star(b)}$; otherwise update $\bar x_k^{\star(b)}$ by \eqref{eq:weighted-average}.
  \EndFor
\EndFor
\State At reporting time, form the observable completed vectors $\widetilde T_{n,b}^\star$ in \eqref{eq:Tstar-completed} if needed.
\If{fixed-$B$ calibration is requested}
  \State Report \eqref{eq:fixedB-CI} or \eqref{eq:fixedB-region}, with their stated fallbacks.
\Else
  \State Report \eqref{eq:basic-CI} or \eqref{eq:intrinsic-region}; use \eqref{eq:wald-region} only
  \Statex \hspace{\algorithmicindent}under the covariance conditions and with its rank/singularity fallback.
\EndIf
\end{algorithmic}
\end{algorithm}

\section{Theoretical Results}\label{sec:theory} 
Work in $T_{x_\star}\M$ with its norms and inner products, and write $x(a)=R_{x_\star}(a)$, $u_k=R_{x_\star}^{-1}(x_k)$ and $u_k^\star=R_{x_\star}^{-1}(x_k^\star)$. Let $\F_k=\sigma(D_1,\ldots,D_k)$ and $\G_k=\F_k\vee\sigma(W_1,\ldots,W_k)$. 
Let $\F_\infty=\sigma(\cup_{k\ge1}\F_k)$. 
Matrix symbols are upright, including coordinate covariance matrices; $H$, $Q$ and $V$ denote basis-free operators on $T_{x_\star}\M$. For $a,b\in T_{x_\star}\M$, $(a\ot b)v=\ip{b}{v}a$. The subscript $\star$ denotes the target, the superscript $\star$ a multiplier path, and a star on probability or law denotes conditioning on $\F_\infty$.
The metric $d_{\BL}$ is the bounded-Lipschitz distance, and
$\rightsquigarrow$ denotes convergence in distribution.

\subsection{Conditions and main guarantees}\label{sec:abstract-assumptions}
The assumptions control the innovation law, accumulated iteration errors, and geometric comparison; the predictable covariance and conditional Lindeberg conditions below are standard martingale central limit conditions \citep[Chapter~3]{HallHeyde1980}, and the \hyperref[app:theory]{theory appendix} supplies sufficient conditions, state bounds and geometric examples.

\begin{assumption}[Innovation and multiplier conditions]\label{ass:innovation}
There are $T_{x_\star}\M$-valued martingale differences $(\xi_k,\F_k)$ and a deterministic positive-semidefinite operator $Q$ such that
\begin{align}
  \frac1n\sum_{k=1}^n
  \Exp(\xi_k\ot\xi_k\mid\F_{k-1})
  &\xrightarrow{p}Q,
  \label{eq:predictable-Q}\\
  \frac1n\sum_{k=1}^n
  \Exp\left[
  \norm{\xi_k}^2\ind\{\norm{\xi_k}>\eps\sqrt n\}
  \mid\F_{k-1}\right]
  &\xrightarrow{p}0
  \quad\text{for every }\eps>0.
  \label{eq:mart-lindeberg}
\end{align}
The multipliers are i.i.d., independent of $\F_\infty$, and satisfy
\begin{equation}\label{eq:weight-second}
  \Exp W=1,
  \qquad \Exp(W-1)^2=1.
\end{equation}
\end{assumption}

\begin{assumption}[Averaged local iterations]\label{ass:abstract-recursion}
There is a nonsingular deterministic linear operator $H:T_{x_\star}\M\to T_{x_\star}\M$ and adapted arrays $b_k,m_k,e_k,m_k^\star,e_k^\star$ such that, on the local chart,
\begin{align}
  u_k
  &=u_{k-1}-\eta_k\{Hu_{k-1}+\xi_k+b_k+m_k+e_k\},
  \label{eq:abstract-base}\\
  u_k^\star
  &=u_{k-1}^\star-\eta_k\{Hu_{k-1}^\star+W_k\xi_k+W_kb_k+m_k^\star+e_k^\star\}.
  \label{eq:abstract-weighted}
\end{align}
Here $b_k$ is $\F_{k-1}$-measurable, $m_k$ is an $(\F_k)$ martingale difference, and $m_k^\star$ is a $(\G_k)$ martingale difference and may include the centered multiplier fluctuation $(W_k-1)Hu_{k-1}^\star$. 
The accumulated conditions are
\begin{align}
  \frac1{\sqrt n}\sum_{k=1}^n b_k&\xrightarrow{p}0,
  &\frac1n\sum_{k=1}^n\norm{b_k}^2&\xrightarrow{p}0,
  \label{eq:bias-abstract}\\
  \frac1n\sum_{k=1}^n
  \Exp(\norm{m_k}^2\mid\F_{k-1})&\xrightarrow{p}0,
  &\frac1n\sum_{k=1}^n
  \Exp(\norm{m_k^\star}^2\mid\G_{k-1})&\xrightarrow{p}0,
  \label{eq:mart-rem-abstract}\\
  \frac1{\sqrt n}\sum_{k=1}^n e_k&\xrightarrow{p}0,
  &\frac1{\sqrt n}\sum_{k=1}^n e_k^\star&\xrightarrow{p}0
  \quad\text{under the joint law}.
  \label{eq:drift-rem-abstract}
\end{align}
Finally,
\begin{equation}\label{eq:state-abstract}
  \sup_{k\ge1}\eta_k^{-1}\Exp\norm{u_k}^2<\infty,
  \qquad
  \sup_{k\ge1}\eta_k^{-1}\Exp\Exp^\star\norm{u_k^\star}^2<\infty.
\end{equation}
Here $\Exp^\star$ denotes expectation over the multiplier stream conditional on $\F_\infty$.
\end{assumption}

\begin{assumption}[Intrinsic averaging and comparison at the estimated base point]\label{ass:intrinsic-comparison}
For $a,b$ in a sufficiently small ball in $T_{x_\star}\M$ and $0\le t\le1$, define
\begin{equation}\label{eq:A-def}
  A_t(a,b)=R_{x_\star}^{-1}
  \left[R_{x(a)}\{tR_{x(a)}^{-1}(x(b))\}\right].
\end{equation}
There is a constant $C$ such that
\begin{equation}\label{eq:A-bound}
  \norm{A_t(a,b)-\{(1-t)a+tb\}}
  \le Ct(\norm a^2+\norm b^2).
\end{equation}
Moreover, with
\begin{equation}\label{eq:Gamma-def}
  \Gamma(a,b)
  =\mathcal P_{x(a)\to x_\star}R_{x(a)}^{-1}\{x(b)\},
\end{equation}
we have
\begin{equation}\label{eq:Gamma-bound}
  \norm{\Gamma(a,b)-(b-a)}
  \le C(\norm a+\norm b)\norm{b-a}.
\end{equation}
The original and weighted states and iterative averages have finite coordinates in this chart under localization and enter a common comparison ball $B_{T_{x_\star}\M}(0,r)$ after an almost surely finite path-specific index. 
The radius $r$ is sufficiently small for the intrinsic averaging bound to yield the $O(k^{-\alpha})$ average rate.
It is permissible for the safeguarded update to differ from its local coordinate formula at finitely many path-specific indices, provided the exact coordinate discrepancies at those indices are included in the arrays $e_k$ and $e_k^\star$ in \eqref{eq:abstract-base}--\eqref{eq:abstract-weighted}.  The reported averages are the full iterative averages in \eqref{eq:base-average} and \eqref{eq:weighted-average}; no common all-path restart is used.
\end{assumption}
These conditions retain familiar requirements of averaged stochastic approximation. \citet[Assumptions A1--A6]{FangXuYang2018} likewise use $\alpha\in(1/2,1)$, population smoothness and a local $L^2$ score modulus, without samplewise differentiability. The state rate in \eqref{eq:state-abstract} and local retraction requirements parallel Assumptions 1--2 of \citet{TripuraneniEtAl2018}. Unlike the samplewise smooth, strongly convex setting of \citet[Assumption 1]{LamWang2026}, our sufficient conditions allow nonsmooth scores and a nonsymmetric population derivative. The additional geometric requirement is joint control of the original and weighted states and full averages in a common chart; this must be verified for each application.

\subsection{Full-path validity}\label{sec:main-theorems}
One sufficient route separates population smoothness from samplewise regularity. A locally $C^{1,\rho}$ population field has remainder $O(\norm u^{1+\rho})$, while the centered stochastic field needs only a vanishing local $L^2$ modulus. Given the state bounds, $\alpha(1+\rho)>1$ makes the accumulated population remainder negligible; for $\rho=1$, $\alpha>1/2$ suffices. Second-order update compatibility controls geometric errors. Full-average localization also needs verification: an invariant contracting averaging basin suffices, whereas tangent-step clipping alone does not ensure a common chart. The \hyperref[sec:primitive]{theory appendix} develops this route.

Choose a normal neighborhood $U_\star$ of $x_\star$ on which transport from $\bar x_n$ to $x_\star$ is unique. Define the completed comparison vectors by
\begin{equation}\label{eq:Tstar-completed}
\begin{aligned}
\widetilde T_n^\star
&=\begin{cases}
\sqrt n R_{\bar x_n}^{-1}(\bar x_n^\star),&\text{if the inverse retraction is available},\\
0_{\bar x_n},&\text{otherwise},
\end{cases}\\
T_n^\star
&=\begin{cases}
\mathcal P_{\bar x_n\to x_\star}\widetilde T_n^\star,&\bar x_n\in U_\star,\\
0_{x_\star},&\text{otherwise}.
\end{cases}
\end{aligned}
\end{equation}
The transport to $x_\star$ is used only in the analysis; reported intervals and regions use the observable differences at $\bar x_n$, and the assumptions make the completion events negligible. The joint expansions remove the state-dependent fluctuations, and the random-base comparison then justifies centering at the observed average.

\begin{theorem}[Joint asymptotic linearity]\label{thm:joint}
Suppose \Cref{ass:innovation,ass:abstract-recursion,ass:intrinsic-comparison} hold with $\eta_k=\eta_0k^{-\alpha}$ and $\alpha\in(1/2,1)$.  Then
\begin{align}
  \sqrt n\,R_{x_\star}^{-1}(\bar x_n)
  &=-H^{-1}\frac1{\sqrt n}\sum_{k=1}^n\xi_k+o_p(1),
  \label{eq:base-linear-new}\\
  \sqrt n\,R_{x_\star}^{-1}(\bar x_n^\star)
  &=-H^{-1}\frac1{\sqrt n}\sum_{k=1}^nW_k\xi_k
    +o_{\Prob^\star}(1)
  \quad\text{in probability}.
  \label{eq:weighted-linear-new}
\end{align}
\end{theorem}

\begin{theorem}[Conditional full-path multiplier validity]\label{thm:bootstrap}
Under the assumptions of \Cref{thm:joint}, let $T_n^\star$ be as in \eqref{eq:Tstar-completed} and $V=H^{-1}Q(H^{-1})^{\dagg}$. Then
\begin{equation}\label{eq:bootstrap-linear-new}
  T_n^\star
  =-H^{-1}\frac1{\sqrt n}\sum_{k=1}^n(W_k-1)\xi_k
   +o_{\Prob^\star}(1)
  \quad\text{in probability},
\end{equation}
and
\begin{equation}\label{eq:bootstrap-BL-new}
  d_{\BL}\{\cL^\star(T_n^\star),N(0,V)\}
  \xrightarrow{p}0.
\end{equation}
Moreover,
\begin{equation}\label{eq:bootstrap-target-new}
  d_{\BL}\left[
  \cL^\star(T_n^\star),
  \cL\{\sqrt nR_{x_\star}^{-1}(\bar x_n)\}
  \right]\xrightarrow{p}0.
\end{equation}
\end{theorem}

Summation by parts reduces each local iteration to its target innovations and negligible accumulated errors. The intrinsic averaging bound makes the iterative manifold average first-order equivalent to its coordinate average, so both averaged coordinates are $O_p(n^{-1/2})$ under the joint law. Substituting them into \eqref{eq:Gamma-bound} makes the inverse-retraction error $O_p(n^{-1})$, negligible after multiplication by $\sqrt n$. Subtracting the linear representations leaves the centered multiplier sum in \eqref{eq:bootstrap-linear-new}. Geometry changes this comparison step, but not the first-order covariance.

\subsection{Fixed-path inference}\label{sec:fixedB-theory}
Put $Z_n=\sqrt nR_{x_\star}^{-1}(\bar x_n)$ and construct $T_{n,b}^\star$ from conditionally independent multiplier paths. Their joint independence in the limit accounts for the uncertainty from using finitely many paths.

\begin{theorem}[Fixed-replication joint limit]\label{thm:fixedB}
Under \Cref{thm:bootstrap}, for every fixed integer $B\ge1$,
\begin{equation}\label{eq:fixedB-joint}
  (Z_n,T_{n,1}^\star,\ldots,T_{n,B}^\star)
  \rightsquigarrow
  (Z_0,Z_1,\ldots,Z_B),
\end{equation}
where $Z_0,\ldots,Z_B$ are independent centered Gaussian tangent vectors, each with covariance $V$.
\end{theorem}

\begin{corollary}[Student and Hotelling-type regions]\label{cor:fixedB}
Suppose \Cref{thm:fixedB} holds.
\begin{enumerate}[label=(\roman*),leftmargin=2.2em]
\item Let $\phi:\M\to\R$ be differentiable at $x_\star$, with differential $D\phi_{x_\star}:T_{x_\star}\M\to\R$, suppose
$\sigma_\phi^2=D\phi_{x_\star}V(D\phi_{x_\star})^\dagg>0$. 
For every fixed $B\ge1$, the interval \eqref{eq:fixedB-CI} satisfies
\begin{equation}\label{eq:fixedB-CI-coverage}
  \Prob\{\phi(x_\star)\in I_{\phi,n,B}(1-\tau)\}\longrightarrow1-\tau.
\end{equation}
\item If $V$ is positive definite and $B\ge d$, the intrinsic region \eqref{eq:fixedB-region} satisfies
\begin{equation}\label{eq:fixedB-region-coverage}
  \Prob\{x_\star\in\mathcal C^{\mathrm{fix}}_{n,B}(1-\tau)\}\longrightarrow1-\tau.
\end{equation}
\end{enumerate}
\end{corollary}

The scalar estimation error is asymptotically independent of $B$ Gaussian bootstrap differences with the same variance, so their root mean square gives a Student-$t_B$ denominator; the degrees of freedom are $B$, not $B-1$, because centering uses the original estimate rather than the empirical bootstrap mean. Fixed-$B$ calibration thus accounts for replication uncertainty without estimating the entire bootstrap distribution.

The $t_B$ penalty can make $B=1$ intervals wide. With growing replication, conditional quantiles remain valid, while covariance estimation additionally requires conditional uniform integrability; the \hyperref[sec:largeB-theory]{theory appendix} gives both guarantees.

\subsection{Approximation bias and its failure boundary}\label{sec:bias}

Approximate gradients, random-direction secants, and smoothed nonsmooth scores can introduce a predictable bias $b_k$; the multiplier paths reproduce the random influence term but not automatically a deterministic root-$n$ shift.

\begin{proposition}[Polynomial approximation-bias regimes and subcritical diagnostic]\label{prop:bias}
For constants $h_0>0$, $r>0$, $\beta>0$, a nonzero $B_0\in T_{x_\star}\M$, and a remainder $\widetilde b_k$, suppose
\begin{equation}\label{eq:bias-polynomial}
  b_k=h_0^r k^{-r\beta}B_0+\widetilde b_k,
  \qquad
  \frac1{\sqrt n}\sum_{k=1}^n\widetilde b_k\xrightarrow{p}0,
  \qquad
  \frac1n\sum_{k=1}^n\norm{\widetilde b_k}^2\xrightarrow{p}0.
\end{equation}
Put $\gamma=r\beta$.  For $\gamma\ge1/2$, suppose in addition that \Cref{ass:innovation,ass:intrinsic-comparison} hold and that all parts of the abstract iteration other than the first condition in \eqref{eq:bias-abstract}, including the state bounds \eqref{eq:state-abstract}, hold with the displayed $b_k$.
\begin{enumerate}[label=(\roman*),leftmargin=2.2em]
\item If $\gamma>1/2$, the approximation bias is negligible at the root-$n$ scale, and the conclusions of \Cref{thm:joint,thm:bootstrap} hold.
\item If $\gamma=1/2$, with $V=H^{-1}Q(H^{-1})^{\dagg}$,
\begin{equation}\label{eq:critical-shift}
  \sqrt nR_{x_\star}^{-1}(\bar x_n)
  \rightsquigarrow
  N\{-2h_0^rH^{-1}B_0,V\}.
\end{equation}
The centered multiplier law still satisfies \eqref{eq:bootstrap-BL-new}, so it does not reproduce the shifted sampling law about $x_\star$. A scalar functional can remain unshifted if its derivative annihilates $H^{-1}B_0$.
\item If $0<\gamma<1/2$ and a bias-aware stability analysis separately justifies the coordinate Abel rearrangement, its deterministic leading term is
\begin{equation}\label{eq:subcritical-shift}
  -\frac{h_0^r}{1-\gamma}n^{1/2-\gamma}H^{-1}B_0\{1+o(1)\},
\end{equation}
whereas the centered weighted bias fluctuation is $o_{\Prob^\star}(1)$.  This coordinate-level calculation alone does not establish the state, intrinsic-average, or random-base remainders for the implemented manifold statistic, and no bootstrap-validity conclusion is asserted in this subcritical regime.
\end{enumerate}
At the boundary, undersmoothing, an analytically justified bias correction, or a multiscale extrapolation that removes the leading $B_0$ term is required for centered inference about $x_\star$.  The subcritical formula is retained only as a diagnostic for the larger deterministic coordinate scale.
\end{proposition}
For smooth Euclidean averaged SGD with exponential weights, the procedure reduces exactly to COnB, see \ref{prop:euclidean-reduction}. On a manifold, the same inference requires localization of the weighted states and full averages and a comparison at the observed base point; \hyperref[sec:primitive]{primitive conditions} and \hyperref[app:geometry]{geometric comparisons} give sufficient routes for these steps. The \hyperref[sec:applications]{geometric applications} specialize the result to spherical means, positive-definite means and medians, and principal subspaces, see \ref{sec:applications}

\section{Numerical Experiments}\label{sec:simulations}
We evaluate the finite-sample coverage of the proposed online multiplier intervals (Proposed; fixed-path Student or 200-path quantile versions) for scalar functionals: first with few bootstrap paths in spherical examples, then across three manifold geometries and with a samplewise nonsmooth score, and finally under different multiplier distributions. All intervals are nominally 95\%, with empirical coverage from 500 independent data streams per configuration. Unless otherwise stated, the iterations use step sizes \(\eta_k=\eta_0k^{-0.7}\) and exponential multiplier weights; design specifications, tuning procedures, and comparator definitions are in the \hyperref[app:simulation-details]{simulation appendix}.

\subsection{Coverage with few bootstrap paths}

To assess fixed-path coverage in small samples, we use two laws on $\mathbb S^2$: a symmetric truncated tangent-normal law with $n\in\{250,500\}$ and an asymmetric four-point law with $n\in\{100,250\}$. Both have intrinsic mean at the north pole $\mathbf e_3$ and scalar estimand $\phi(x)=x_1$. Independent pilots selected $\eta_0=0.8$ and $1.0$, respectively, by base-estimator mean squared geodesic error before the 500-stream evaluations.

\Cref{tab:sphere-small-sample} compares Proposed Student intervals, Proposed (quantile), a tangent sandwich interval and a matched offline basic bootstrap with 200 trajectory reruns. Methods share observations, initialization, tuning and the base estimate; all paths average from the first update without burn-in. The offline method replays the same iteration on each resampled ordered dataset.
\begin{table}[!htb]
\centering
\caption{Nominal 95\% coverage in localized spherical designs.
Proposed Student and quantile intervals use \eqref{eq:fixedB-CI} and
\eqref{eq:basic-CI}; the quantile and offline methods each use 200 replicates.}
\label{tab:sphere-small-sample}
\begin{tabular}{lrrrr}
\toprule
 & \multicolumn{2}{c}{Symmetric} & \multicolumn{2}{c}{Skew}\\
Method & $n=250$ & $n=500$ & $n=100$ & $n=250$\\
\midrule
Proposed ($B=1$) & 0.940 & 0.938 & 0.954 & 0.950\\
Proposed ($B=3$) & 0.956 & 0.954 & 0.942 & 0.938\\
Proposed ($B=5$) & 0.960 & 0.950 & 0.938 & 0.942\\
Proposed ($B=10$) & 0.952 & 0.946 & 0.936 & 0.932\\
Proposed (quantile) & 0.968 & 0.944 & 0.932 & 0.924\\
Sandwich & 0.942 & 0.930 & 0.910 & 0.910\\
Offline & 0.904 & 0.926 & 0.840 & 0.878\\
\bottomrule
\end{tabular}
\end{table}

With only three to ten multiplier paths, the proposed method achieves coverage near nominal and improves on offline replay across all four settings. The gains are most pronounced under the skew law at $n=100$, highlighting the effectiveness of few-path inference even in a curved, asymmetric problem with a small sample. Coverage is also broadly comparable to that of sandwich and 200-path quantile intervals.
The improved calibration relative to offline replay comes with wider intervals, although their lengths decrease substantially as $B$ increases; full length comparisons are in the simulation appendix.

A wider-cap comparison uses $n=1000$ observations from a von Mises--Fisher law on $\mathbb S^2$, with scalar estimand $\phi(x)=x_1$.  
\Cref{tab:sphere-comparator} shows coverage near nominal for Proposed Student and sandwich intervals, versus 0.900 for offline replay.

\begin{table}[!htb]
\centering
\caption{Nominal 95\% coverage in the wider-cap spherical design ($q=2$, $n=1000$). Proposed (quantile) and offline basic intervals each use 200 replicates.}
\label{tab:sphere-comparator}
\begin{tabular}{rrrrrrr}
\toprule
\multicolumn{4}{c}{Proposed (Student)} & \shortstack{Proposed\\(quantile)} & Sandwich & Offline\\
$B=1$ & $B=3$ & $B=5$ & $B=10$ & $B=200$ & & $B=200$\\
\midrule
0.948 & 0.958 & 0.950 & 0.970 & 0.964 & 0.948 & 0.900\\
\bottomrule
\end{tabular}
\end{table}

\subsection{Coverage across geometries and nonsmooth scores}
We next consider spherical means, affine-invariant positive-definite matrix means, and principal subspaces on Grassmann manifolds. Their scalar functionals are, respectively, the coordinate $x_1$, the largest log eigenvalue at a simple-eigenvalue matrix target, and the projector entry $(\mathrm{U}\mathrm{U}^{\trans})_{r,r+1}$, where $\mathrm U$ is an orthonormal basis of the principal subspace. Dimensions and distributional parameters are specified in the simulation appendix.

\begin{figure}[!htb]
\centering
\includegraphics[width=0.96\textwidth]{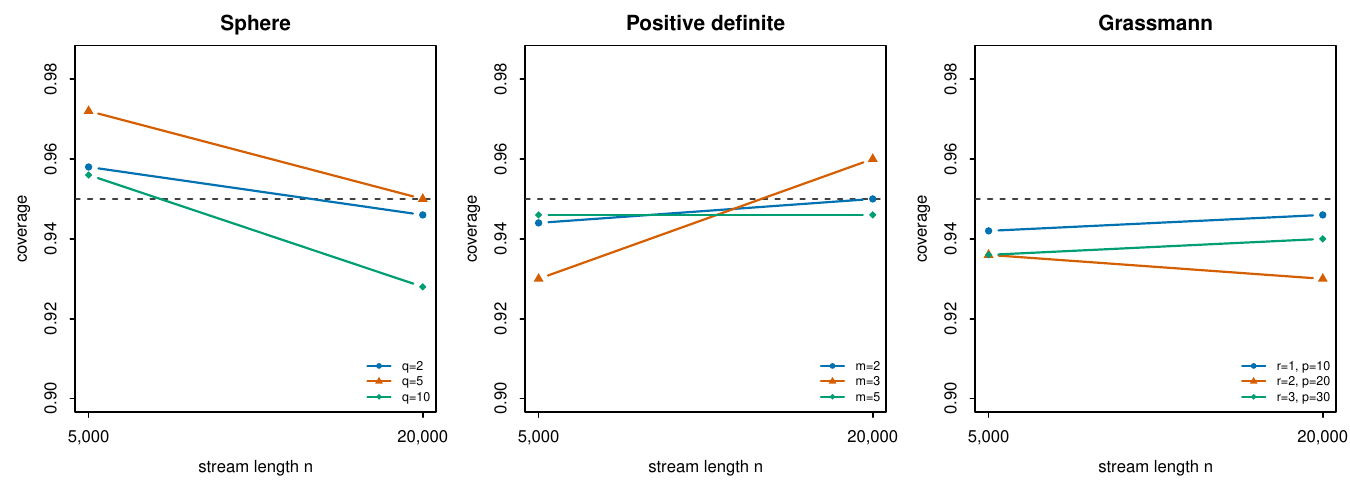}
\caption{Proposed ($B=10$) coverage in regular designs; the dashed line marks 95\%. Spherical laws differ between horizons, so connecting segments are visual guides rather than same-law sample-size trends.}
\label{fig:formal-coverage}
\end{figure}

With $B=10$, the Proposed intervals achieve coverage near nominal for spherical means, affine-invariant positive-definite means and Grassmann principal subspaces at $n\in\{5000,20000\}$, without estimating population derivatives (\Cref{fig:formal-coverage}).

To assess a samplewise nonsmooth score, we compare the positive-definite Fr\'echet mean and geometric median. In the specified Gaussian model with sign-symmetric contamination, affine-invariant reflection symmetry and uniqueness preserve their common target. 
For $3\times3$ positive-definite matrices at $n=5000$, \Cref{tab:spd-robustness} shows similar coverage near nominal at contamination probabilities $0$, $0.05$ and $0.15$.

\begin{table}[!htb]
\centering
\caption{Nominal 95\% coverage of Proposed ($B=10$) for the largest log eigenvalue of the common positive-definite mean and median ($3\times3$ matrices, $n=5000$).}
\label{tab:spd-robustness}
\begin{tabular}{lrr}
\toprule
Contamination probability & Fr\'echet mean & Geometric median\\
\midrule
0    & 0.930 & 0.936\\
0.05 & 0.952 & 0.948\\
0.15 & 0.952 & 0.956\\
\bottomrule
\end{tabular}
\end{table}

\subsection{Sensitivity to the multiplier distribution}

A separate cap design at $n=5000$ compares weights on identical streams, with concentration 6, radius 1.20, start 0.45 radians from the north pole, $\eta_0=0.6$ and estimand $\phi(x)=x_1$ as in the wider-cap comparison. Exponential, Poisson and signed two-point weights give Proposed ($B=10$) coverage of 0.946, 0.954 and 0.964. The signed law assigns probabilities 0.2 and 0.8 to $-1$ and $1.5$, giving mean and variance one. All three laws therefore preserve the target innovation covariance in the first-order multiplier representation. \tfill{Conditionally on the original stream, the multiplier sum $n^{-1/2}\sum_{k=1}^n(W_k-1)\xi_k$ has mean zero and covariance $n^{-1}\sum_{k=1}^n\xi_k\ot\xi_k$. Only the unit weight variance enters this covariance, whose convergence in probability to $Q$ follows from \Cref{ass:innovation}.} They need not produce identical finite-sample paths: Poisson weights can omit an update, while a negative weight reverses its direction. The comparable coverage illustrates similar behavior under these distinct perturbations, without showing superiority of signed weights. Their use still requires the weighted stability and localization conditions of the theorem. Implementation details are in the \hyperref[app:simulation-details]{simulation appendix}; the \hyperref[app:ibtracs]{data appendix} gives descriptive inference for 738 North Atlantic storm locations.

\section{Concluding Remarks}\label{sec:conclusion}

Full-path multipliers provide online inference for Riemannian Polyak-Ruppert averages without plug-in derivative or covariance estimation. The proposed construction supports valid fixed-path Student intervals and intrinsic regions with few multiplier paths, while growing replication yields conditional quantiles. The random-base comparison resolves the difficulty of comparing tangent spaces and allows the same weighted trajectories to be used directly for manifold-valued inference. The framework also accommodates samplewise nonsmooth scores and identifies when accumulated approximation bias remains negligible or shifts the limiting law.

The guarantees require local stability and control of accumulated approximation bias in fixed dimension. Dependence-aware weights and bias-corrected oracles are natural extensions beyond the present martingale and negligible-bias conditions.

\clearpage

\section*{AI use statement}
Generative AI tools were used to revise wording and improve the clarity and readability of the manuscript; to identify potentially relevant literature, references, and search directions, and to assist with research execution, including drafting, debugging, and refining code. For mathematical proofs, AI was used primarily to review and check the authors’ arguments for potential errors, inconsistencies, or gaps. AI-assisted manuscript revisions, literature-search suggestions, feedback on research ideas and proofs, and generated or modified code were reviewed and verified by the authors before incorporation into the final work. The authors independently assessed the accuracy and appropriateness of these outputs and retain full responsibility for the methodology, proofs, implementation, analysis, and conclusions presented in the paper.

\section*{Reproducibility Statement}
The code and an accompanying \texttt{README} file are submitted with the
manuscript. We provide a dedicated reproduction script for each figure and
table, allowing the reported results to be reproduced individually. The
\texttt{README} provides instructions for running the code.

\clearpage
\appendix
\section{Simulation Details and Additional Results}\label{app:simulation-details}

This appendix specifies the experiments in the main text and reports additional comparisons for spherical means. Coverage uses all 500 streams in each configuration; estimation error is the median geodesic distance of the localized spherical estimator from its population target.

\subsection{Protocol and implementation}\label{sec:sim-common}

All iterations use $\alpha=0.7$. The step constants are $\eta_0=0.6$ on the original sphere designs, $0.8$ and $1.0$ on the symmetric and skew localized sphere designs, $0.3$ for positive-definite means, $0.2$ for positive-definite medians, and $0.1$ on Grassmann manifolds.
The proposed fixed-path intervals are evaluated with $B\in\{1,3,5,10\}$ replicas, using exponential weights and the Student calibration in \eqref{eq:fixedB-CI}.
We additionally compare Proposed (quantile) with $B=200$, a tangent plug-in sandwich interval, and an offline trajectory bootstrap with 200 resampled trajectories.
Methods compared within a configuration receive the same original stream and base fit; the two 200-replicate procedures use equal replication budgets. The offline comparator uses basic quantiles, not the fixed-replication Student calibration of COnB \citep{LamWang2026}. Full regions are formed only when $B$ is at least the tangent dimension.

The sphere designs use cap-supported von Mises--Fisher laws and rotational mixtures in dimensions $q\in\{2,5,10\}$ and the scalar functional $\phi(x)=\mathbf e_1^{\trans}x=x_1$, whose target is zero at the north pole. Positive-definite observations have the form $\mathrm{Y}=\mathrm{A}^{1/2}\exp(\mathrm{E})\mathrm{A}^{1/2}$ for $m\in\{2,3,5\}$, with sign-symmetric Gaussian or contaminated $\mathrm{E}$. The target $\mathrm{A}$ has a simple largest eigenvalue, so its largest-log-eigenvalue functional satisfies \eqref{eq:logmax-derivative}. Grassmann designs use $(r,p)\in\{(1,10),(2,20),(3,30)\}$ and Gaussian observations with eigengap 0.6. Their scalar target is the smooth projector entry $(\mathrm{U}\mathrm{U}^{\trans})_{r,r+1}$. The cross-geometry comparison uses $n\in\{5000,20000\}$.

The iterations average all post-update states from $k=1$ without burn-in. In the two localized spherical families, independent pilots selected a common step constant for every method by minimizing pooled mean squared geodesic error of the base average, not by optimizing coverage. The symmetric pilot used 40 streams per $(n,\eta_0)$ cell and $\eta_0\in\{0.4,0.6,0.8\}$; the skew pilot used 50 and $\eta_0\in\{0.4,0.6,0.8,1.0\}$. The selected constants were fixed before the 500-stream evaluations, whose seeds are disjoint from the pilots. All reported coverage denominators include all 500 streams. Coverage differences are interpreted together with their paired Monte Carlo variability; small numerical differences do not establish an ordering of methods.

\subsection{Design details}
\paragraph{Sphere designs.}
Write $\mathbf e_{q+1}$ for the north pole.  In a cap design, $T=\ip{Y}{\mathbf e_{q+1}}$ has density proportional to
\[
  \exp(\kappa t)(1-t^2)^{(q-2)/2},
  \qquad \cos(r_{\mathrm{cap}})\le t\le1,
\]
and, conditional on $T$, the direction in $\mathbf e_{q+1}^{\perp}$ is uniform.  A rotational mixture draws two components with equal probability after rotating them by $\pm0.25$ radians in the first--last coordinate plane; the component cap radius is $r_{\mathrm{cap}}-0.25$.  The primary $(n,\text{design},\kappa,r_{\mathrm{cap}})$ settings are $(5000,\text{cap},8,1.45)$ and $(20000,\text{mixture},6,1.10)$ for $q=2$; $(5000,\text{cap},8,1.20)$ and $(20000,\text{mixture},6,1.10)$ for $q=5$; and $(5000,\text{mixture},8,1.20)$ and $(20000,\text{cap},6,1.10)$ for $q=10$.  The representative $q=2$, $n=1000$ cell in \Cref{tab:sphere-comparator} is a cap design with $(\kappa,r_{\mathrm{cap}})=(4,1.20)$; the weight comparison uses $(q,n,\kappa,r_{\mathrm{cap}})=(2,5000,6,1.20)$.  Reported cells start from $\Expmap_{\mathbf e_{q+1}}(0.45 \mathbf e_1)$, use the exponential map and logarithm, and rescale steps above 0.75.

\paragraph{Localized small-sample sphere designs.}
Both laws take $Y=\Expmap_{\mathbf e_3}(v)$ for a two-dimensional tangent vector $v$. In the symmetric design, $v\sim N(0,0.3^2\mathrm{I}_2)$ conditional on $\norm v\le0.70$. In the skew design, its support is
\[
 (0.55,0.05),\quad(-0.35,0.45),\quad(-0.45,-0.25),\quad(-17/60,-0.55),
\]
with probabilities $0.40,0.25,0.20,0.15$, respectively. The support radii are 0.70 and approximately 0.619 radians. Both tangent means are zero; together with strong convexity on the supporting cap, this identifies the same intrinsic target without imposing symmetry on the second law.

The symmetric configurations use $n=250,500$, start at $\Expmap_{\mathbf e_3}(0.30\mathbf e_1)$, take $\eta_0=0.8$, and rescale proposed tangent steps above 0.60. The skew configurations use $n=100,250$, start at $\Expmap_{\mathbf e_3}(0.25\mathbf e_1)$, take $\eta_0=1.0$, and rescale steps above 0.55. In both designs, the observations lie in a cap of radius below $\pi/4$. Applying \Cref{cor:sphere-averaging-basin} additionally requires the original and weighted states to remain in such a common cap, converge to the target, and satisfy the stated rates; tangent-step rescaling alone does not certify these path conditions. A single 200-path multiplier run supplies the nested fixed-$B$ and quantile intervals, and each configuration also uses 200 offline trajectory resamples.

The spherical score is $\Psi(x;Y)=-\Log_x(Y)$. Each path moves toward the incoming observation along a geodesic, with the multiplier changing its step before tangent-step rescaling. Its iterative average also uses the exponential map and logarithm, rather than normalizing an arithmetic average of the observations.

\paragraph{Interval lengths.}
\Cref{tab:sphere-lengths} reports median lengths of the 95\% intervals in the localized and wider-cap comparisons. Length is the upper endpoint minus the lower endpoint, in units of $\phi(x)=x_1$, without truncation to its range $[-1,1]$. Each median uses all 500 streams. The long $B=1$ intervals reflect the Student critical value with one degree of freedom; using three to ten paths substantially reduces this cost.

\begin{table}[!htb]
\centering
\caption{Median 95\% interval lengths for the spherical coverage comparisons. Proposed (quantile) and offline methods each use 200 replicates; all methods share the base estimator within a configuration.}
\label{tab:sphere-lengths}
\begin{tabular}{lrrrrr}
\toprule
 & \multicolumn{2}{c}{Symmetric} & \multicolumn{2}{c}{Skew} & Wider cap\\
Method & $n=250$ & $n=500$ & $n=100$ & $n=250$ & $n=1000$\\
\midrule
Proposed ($B=1$) & 0.3280 & 0.2316 & 0.8337 & 0.4994 & 0.2814\\
Proposed ($B=3$) & 0.1075 & 0.0720 & 0.2782 & 0.1775 & 0.0963\\
Proposed ($B=5$) & 0.0927 & 0.0634 & 0.2384 & 0.1515 & 0.0818\\
Proposed ($B=10$) & 0.0841 & 0.0573 & 0.2217 & 0.1370 & 0.0737\\
Proposed (quantile) & 0.0764 & 0.0526 & 0.1996 & 0.1256 & 0.0662\\
Sandwich & 0.0684 & 0.0484 & 0.1832 & 0.1162 & 0.0609\\
Offline & 0.0740 & 0.0516 & 0.1966 & 0.1232 & 0.0652\\
\bottomrule
\end{tabular}
\end{table}

\paragraph{Representative spherical comparators.}
At $\bar x_n$, let the columns of $\mathrm{E}$ be an orthonormal basis of $T_{\bar x_n}\mathbb S^q$, set $z_i=\mathrm{E}^{\trans}\Log_{\bar x_n}(Y_i)$ and $r_i=\norm{z_i}$, and put
\[
 \widehat {\mathrm{H}}=\frac1n\sum_{i=1}^n
 \left[\nu_i\nu_i^{\trans}+\frac{r_i}{\tan r_i}
 \{\mathrm{I}-\nu_i\nu_i^{\trans}\}\right],
 \qquad
 \widehat {\mathrm{Q}}=\frac1n\sum_{i=1}^n(z_i-\bar z)(z_i-\bar z)^{\trans},
\]
where $\bar z=n^{-1}\sum_{i=1}^nz_{i}$, $\nu_i=z_i/r_i$ when $r_i>0$ and the corresponding Hessian summand is $\mathrm{I}$ when $r_i=0$.  The plug-in sandwich is $\widehat {\mathrm{V}}=\widehat {\mathrm{H}}^{-1}\widehat {\mathrm{Q}}(\widehat {\mathrm{H}}^{-1})^{\trans}$, and its interval for $\phi(x)=x_1$ uses the standard error $\{(\mathrm{E}^{\trans} \mathbf e_1)^{\trans}\widehat {\mathrm{V}}(\mathrm{E}^{\trans}\mathbf e_1)/n\}^{1/2}$ and the standard-normal critical value.  For each of the 200 offline trajectory replicates, $n$ indices are drawn independently with replacement, the base iteration and its full iterative average are replayed on the resulting ordered sample with the same initialization, step schedule, retraction, and step-rescaling rule, and the first coordinate is recorded.  The reported offline interval is the basic interval formed from the type-8 empirical 0.025 and 0.975 quantiles of those replicate-minus-original differences.

\paragraph{Positive-definite designs.}
For $m\in\{2,3,5\}$, let
\[
 \mathrm{A}=\operatorname{diag}\{\exp(\ell_1),\ldots,\exp(\ell_m)\},
 \qquad
\ell_j=-0.3+\frac{0.6(j-1)}{m-1}, \qquad j=1,\ldots,m,
\]
and set $\mathrm{Y}=\mathrm{A}^{1/2}\exp(\mathrm{E})\mathrm{A}^{1/2}$.  The independent diagonal entries of the symmetric Gaussian matrix $\mathrm{E}$ have variance $0.25^2$, and its independent upper-triangular entries have variance $0.25^2/2$.  In the robustness cells, with probability $\varepsilon\in\{0,0.05,0.15\}$ we add
\[S\,1.2\,\operatorname{diag}(c_1,\ldots,c_m),
 \qquad \Prob(S=1)=\Prob(S=-1)=1/2,
\]
to $\mathrm{E}$, where $c_j=-1+\frac{2(j-1)}{m-1},\qquad j=1,\ldots,m$.  Primary mean cells use $n\in\{5000,20000\}$ for each $m$; the mean--median comparison uses $m=3$, $n=5000$.  All paths start at $\mathrm{I}_m$, use the affine-invariant exponential and logarithm, and rescale steps above 0.8.

\paragraph{Grassmann designs.}
For $(r,p)\in\{(1,10),(2,20),(3,30)\}$, the target is the span of the first $r$ coordinate vectors.
Gaussian observations have independent coordinates with covariance eigenvalues. Specifically,  for $r=2$, set $\lambda_1=2.15$; for $r=3$, set
$(\lambda_1,\lambda_2)=(2.30,2.00)$.
In all cases, $\lambda_r=1+\delta,\qquad \lambda_{r+1:p}=1$. 
The regular cells use $\delta=0.6$ and $n\in\{5000,20000\}$. Initialization rotates the last active target vector toward the first inactive coordinate by 0.10 radians. Updates use the QR retraction and graph inverse in \Cref{lem:implementation-geometry}, with step rescaling above 0.8.

\section{North Atlantic Tropical-Cyclone Locations}\label{app:ibtracs}

\subsection{Data, estimand, and preprocessing}\label{sec:ibtracs-data}

The International Best Track Archive for Climate Stewardship (IBTrACS) reconciles tropical-cyclone best-track records from multiple agencies \citep{KnappEtAl2010}.  We use the publicly accessible official NOAA/NCEI North Atlantic subset, version 4r01, downloaded on 21 August 2026 \citep{NOAAIBTrACS}.

The spatial location at which a storm first reaches a tropical-storm threshold is itself a scientifically studied summary: for example, \citet{WangEtAl2016Genesis} use such locations to study meridional variability across the North Atlantic and eastern North Pacific.  Our selection below is deliberately more conservative in its labeling.  It is defined by the first qualifying IBTrACS archive category and includes subtropical records, so we call it a first recorded tropical/subtropical location rather than physical genesis.

We first restrict to complete seasons 1980--2025, North Atlantic main-track records with finite coordinates, and archive times at which the combined \texttt{NATURE} field classifies a storm as tropical or subtropical (\texttt{TS} or \texttt{SS}).  Within each storm identifier we then retain the earliest qualifying archive time, and finally order the storm-level observations chronologically.  The retained point is therefore the first \emph{recorded} tropical or subtropical location in the archive, not a physical genesis location.  The resulting dataset contains 738 storms, of which 89.3\% first enter the retained set as tropical.

Locations are represented as unit vectors on $\mathbb S^2$, and the estimand is the intrinsic center of the pooled storm-level distribution.  Every retained point is less than $\pi/2$ from the reported center. This describes the observed spatial configuration, but does not certify common-chart membership of the original and weighted trajectories or their full averages. We report descriptive intervals under a working-independence model for the chronological archive; robustness to serial dependence or distributional change is not asserted.

\subsection{Online analysis and uncertainty}\label{sec:ibtracs-results}

We process storms chronologically from the north pole using the spherical exponential map, $\eta_k=0.6k^{-0.7}$, and exponential weights.  Proposed tangent steps are rescaled only when their norm exceeds 0.75.  The first ten weighted paths give fixed-$B=10$ intervals and all 200 paths give large-$B$ basic intervals.  Longitude is unwrapped locally around the estimate, so longitude and latitude are smooth scalar coordinates on the observed cap. The reported intervals are marginal for each coordinate, not a simultaneous two-coordinate confidence region.

\begin{table}[!htb]
\centering
\caption{North Atlantic first recorded tropical/subtropical locations, 1980--2025.  Coordinates and interval endpoints are in degrees; west longitudes are negative.  All displayed intervals are descriptive working-independence multiplier intervals.}
\label{tab:ibtracs}
\begin{tabular}{lrrr}
\toprule
Quantity & Estimate & 95\% fixed $B=10$ & 95\% large $B=200$\\
\midrule
Longitude & $-59.88$ & $[-61.77,-57.98]$ & $[-61.57,-58.37]$\\
Latitude  & $ 21.92$ & $[ 21.06, 22.78]$ & $[ 21.11, 22.49]$\\
\bottomrule
\end{tabular}
\end{table}

The chronological online center is $59.88^\circ$W, $21.92^\circ$N.  The fixed-$B=10$ interval places its longitude between $61.77^\circ$W and $57.98^\circ$W and its latitude between $21.06^\circ$N and $22.78^\circ$N.  The large-$B=200$ intervals are similar, locating the pooled center in the tropical western Atlantic with closely agreeing uncertainty summaries.

\begin{figure}[!htb]
\centering
\includegraphics[width=0.82\textwidth]{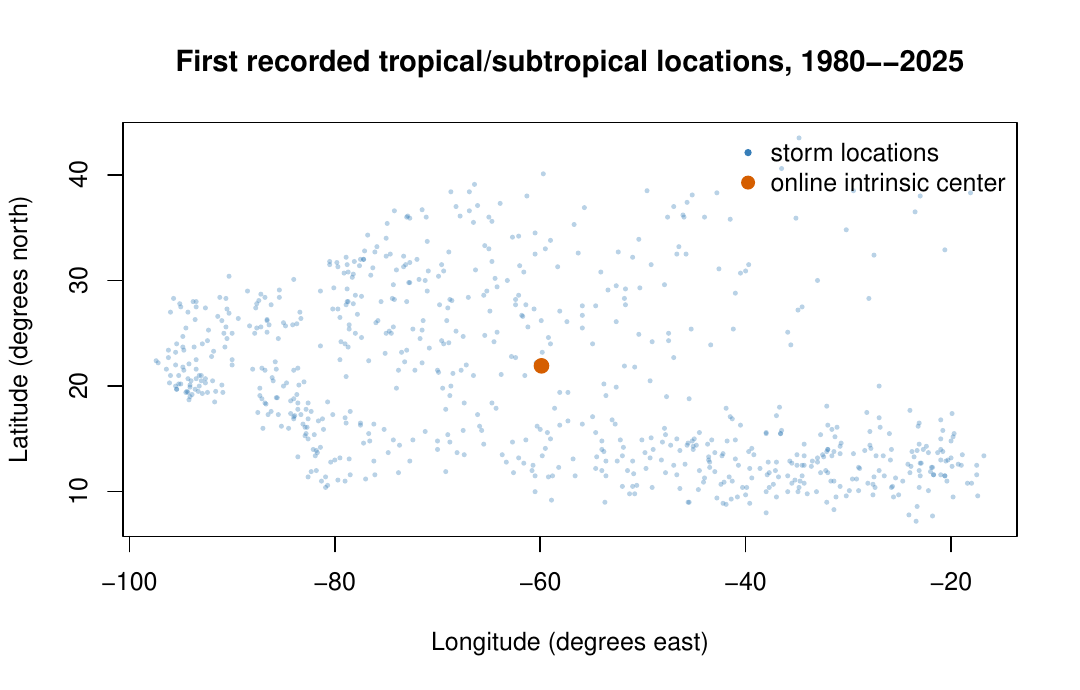}
\caption{Retained storm-level locations and the chronological online intrinsic center.  The plotted points are first archive records satisfying the tropical/subtropical filter, not physical genesis estimates.}
\label{fig:ibtracs}
\end{figure}
\clearpage
\section{Theory Supplement}\label{app:theory}
\subsection{Calibration and implementation details}\label{app:implementation}
The theorem permits signed weights.  Examples satisfying \eqref{eq:weight-normalization} are $W\sim\operatorname{Exp}(1)$, $W\sim\operatorname{Poisson}(1)$, $W=1+\epsilon$ with $\epsilon\in\{-1,1\}$ equiprobably, and $W=1+G$ with $G\sim N(0,1)$.  Nonnegative weights may reduce finite-sample path instability, but nonnegativity is not needed for first-order validity once the weighted state bound is verified.

\subsubsection{Two calibration regimes}\label{sec:outputs}

Fix $0<\tau<1$ as the confidence error probability. For tangent vectors, $(a\ot b)v=\ip{b}{v}a$ denotes the rank-one operator. The completed vectors are defined in \eqref{eq:Tstar-completed}. On $\{\bar x_{n}\in U_{\star}\}$, which depends only on the original stream, the same isometry preserves inner products for every replicate. The fixed-space representation is used only for analysis; the algorithm does not use $x_{\star}$.

\paragraph{Fixed-replication calibration.}
Let $B\ge1$ be fixed and let $\bar x_n^{\star(1)},\ldots,\bar x_n^{\star(B)}$ be conditionally independent multiplier averages.  For a scalar functional $\phi:\M\to\R$, use \eqref{eq:fixedB-scale}--\eqref{eq:fixedB-CI}, where $t_{B,1-\tau/2}$ is the Student-$t$ quantile with $B$ degrees of freedom.  The denominator is centered at the observed original estimate, so the degrees of freedom are $B$, not $B-1$.
If $s_{\phi,n,B}=0$, define $I_{\phi,n,B}(1-\tau)=\R$; under the positive limiting-variance condition in \Cref{cor:fixedB}, this fallback event has probability tending to zero.

For a full intrinsic region, let $\widetilde T_{n,b}^\star$ be the completed statistic in \eqref{eq:Tstar-completed} for path $b$ and put
\begin{align*}
  \widehat S_{n,B}&=\frac1B\sum_{b=1}^B
  \widetilde T_{n,b}^\star\ot\widetilde T_{n,b}^\star.
\end{align*}
When $B\ge d$ and $\widehat S_{n,B}$ is invertible, restrict attention to the retractive neighborhood of $\bar x_n$ on which $R_{\bar x_n}^{-1}$ is single valued and define
\begin{equation}\label{eq:fixedB-region}
  \mathcal C^{\mathrm{fix}}_{n,B}(1-\tau)
  =\left\{x:\
  n\ip{R_{\bar x_n}^{-1}(x)}{\widehat S_{n,B}^{-1}R_{\bar x_n}^{-1}(x)}_{\bar x_n}
  \le c_{d,B,\tau}\right\},
\end{equation}
with
\begin{equation}\label{eq:fixedB-critical}
  c_{d,B,\tau}
  =\frac{dB}{B-d+1}F_{d,B-d+1;1-\tau},
\end{equation}
where $F_{a,b;\gamma}$ denotes the $\gamma$ quantile of the $F$ distribution with $(a,b)$ degrees of freedom.  For $d=1$, the critical-value identity $F_{1,B;1-\tau}=t_{B,1-\tau/2}^2$ makes \eqref{eq:fixedB-region} coincide with \eqref{eq:fixedB-CI} for the Euclidean identity functional, or for the one-dimensional local coordinate used to define the intrinsic region.  For a nonlinear scalar functional, the two procedures instead have the same first-order pivotal calibration through the delta method.
If $\widehat S_{n,B}$ is singular, define $\mathcal C^{\mathrm{fix}}_{n,B}(1-\tau)=\M$.  Positive-definite $V$ and $B\ge d$ make this fallback event asymptotically negligible.  The borderline choice $B=d$ is valid but has one denominator degree of freedom and is consequently not a practical default.

\paragraph{Large-replication calibration.}
When $B=B_n\to\infty$, empirical bootstrap quantiles approximate the full conditional law.  Let $q_\gamma^\star$ be the empirical $\gamma$ quantile of
\[
  \sqrt n\{\phi(\bar x_n^{\star(b)})-\phi(\bar x_n)\},
  \qquad b=1,\ldots,B_n.
\]
The basic interval is
\begin{equation}\label{eq:basic-CI}
  \left[
  \phi(\bar x_n)-\frac{q_{1-\tau/2}^\star}{\sqrt n},
  \quad
  \phi(\bar x_n)-\frac{q_{\tau/2}^\star}{\sqrt n}
  \right].
\end{equation}
If $c_{1-\tau}^\star$ is the empirical $(1-\tau)$ quantile of
$\norm{\widetilde T_{n,b}^\star}_{\bar x_n}$, the norm region is
\begin{equation}\label{eq:intrinsic-region}
  \mathcal C_{n,1-\tau}
  =\left\{R_{\bar x_n}(v/\sqrt n):
     v\in T_{\bar x_n}\M,\ \norm v_{\bar x_n}\le c_{1-\tau}^\star\right\}.
\end{equation}
The set in \eqref{eq:intrinsic-region} includes only vectors for which $v/\sqrt n$ lies in the domain of $R_{\bar x_{n}}$. A large-$B$ bootstrap covariance estimator is
\begin{equation}\label{eq:Vhat}
  \widehat V_{n,B}^\star
  =\frac1B\sum_{b=1}^B
  (\widetilde T_{n,b}^\star-\overline T_{n,B}^\star)
  \ot
  (\widetilde T_{n,b}^\star-\overline T_{n,B}^\star),
  \qquad
  \overline T_{n,B}^\star=\frac1B\sum_{b=1}^B\widetilde T_{n,b}^\star.
\end{equation}
If the estimator in \eqref{eq:Vhat} is invertible, restrict the corresponding Wald region to a retractive neighborhood $U_{\bar x_{n}}$ on which the inverse is single-valued:
\begin{equation}\label{eq:wald-region}
  \left\{x\in U_{\bar x_{n}}:
   n\ip{R_{\bar x_n}^{-1}(x)}{
   (\widehat V_{n,B}^\star)^{-1}R_{\bar x_n}^{-1}(x)}_{\bar x_n}
   \le \chi^2_{d,1-\tau}\right\}.
\end{equation}
Because the $B$ centered vectors in \eqref{eq:Vhat} have rank at most $B-1$, invertibility requires $B\ge d+1$.  If this rank condition fails or $\widehat V_{n,B}^\star$ is singular, define the Wald region to be $\M$.  Under the positive-definite limit in \Cref{prop:covariance} and $B_n\to\infty$, this fallback event has probability tending to zero.
Fixed-$B$ calibration and large-$B$ conditional-law approximation are different uses of the same paths.  The former gives asymptotic coverage with very few paths but uses Gaussian pivotality; the latter estimates the conditional shape of the bootstrap law and requires Monte Carlo growth.

\subsubsection{Implementation details}\label{sec:algorithm}

All inverse maps used by the iterative averages must be defined along the implemented paths; the reporting completion does not define a missing earlier update. When a safeguard is used, it is part of each state update. For a state representation using $O(d)$ numbers, the additional path storage is $O(Bd)$ and every observation requires $B$ extra oracle evaluations and retractions.  A Grassmann QR representation instead stores $p\times r$ matrices, and forming a full reporting covariance requires $O(d^{2})$ additional storage. The data point and base seed can be discarded after all paths have been updated.  An offline replay using the same ordered data, seeds, weights, localization rule, and arithmetic produces the same paths; the method does not rely on resampling stored observations. For a growing replication count $B_{n}$, all paths for a chosen horizon are initialized before processing its stream; adding a path after discarding earlier data does not reproduce a full-history path.

\subsubsection{Fixed-coordinate notation and probability conventions}\label{sec:notation}

Fix a relatively compact retractive neighborhood $U$ of $x_\star$ and write
\[
  x(a)=R_{x_\star}(a),
  \qquad u_k=R_{x_\star}^{-1}(x_k),
  \qquad u_k^\star=R_{x_\star}^{-1}(x_k^\star).
\]
All unlabelled norms and inner products are those of $T_{x_\star}\M$.  Let
\[
  \F_k=\sigma(D_1,\ldots,D_k),
  \qquad
  \G_k=\F_k\vee\sigma(W_1,\ldots,W_k).
\]
Write $\F_{\infty}=\sigma(\bigcup_{k\ge1}\F_{k})$; the starred probability and expectation operators are defined in the main text.

For random variables $Y_n^\star$ depending on both streams, $Y_n^\star=o_{\Prob^\star}(1)$ in probability means that, for every $\eps>0$,
\[
  \Prob\{\Prob^\star(\norm{Y_n^\star}>\eps)>\eps\}\longrightarrow0.
\]
For probability measures on the finite-dimensional space $T_{x_\star}\M$, $d_{\BL}$ denotes the bounded-Lipschitz distance, and $N(0,V)$ denotes the centered Gaussian law with covariance $V$. Matrix subscripts $F$ indicate the Frobenius inner product or norm; $\vecop$ stacks matrix columns and $I$ is the identity of the relevant space.  For a linear operator $A$, $A^\dagg$ denotes its adjoint. Upright symbols denote matrices; tangent vectors and basis-free linear operators retain their intrinsic notation.

\subsection{Additional discussions}\label{app:conditions}
\paragraph{Moment conditions }\label{rem:moment-sufficient}
Equations \eqref{eq:predictable-Q}--\eqref{eq:mart-lindeberg} are the covariance and conditional Lindeberg assumptions of the martingale central limit theorem \citep{HallHeyde1980}.  A uniform $2+\delta$ moment is sufficient for \eqref{eq:mart-lindeberg} but not necessary, while \eqref{eq:predictable-Q} must be verified separately.  Likewise, \eqref{eq:weight-second} is enough for the conditional multiplier central limit theorem.  Higher multiplier moments or nonnegativity are stability conveniences rather than requirements for weak convergence.

\paragraph{Approximation bias.}\label{rem:weighted-bias}
The full oracle is multiplied by $W_k$, so a predictable approximation bias appears as $W_kb_k$.  The decomposition
\[
  \frac1{\sqrt n}\sum_{k=1}^nW_kb_k
  =\frac1{\sqrt n}\sum_{k=1}^nb_k
   +\frac1{\sqrt n}\sum_{k=1}^n(W_k-1)b_k
\]
explains the two conditions in \eqref{eq:bias-abstract}.  The vector-sum condition allows directional cancellation; the conditional variance of the second term is $n^{-1}\sum\norm{b_k}^2$.

\paragraph{Localization.}\label{rem:localization}
A safeguard, when used, is part of the implemented state iteration. Localization does not restart or replace its reported full iterative averages. Eventual inactivity means that the ordinary update is applied at the current safeguarded state; it does not identify this path with a separately run unsafeguarded path having a different history.  The assumption above concerns both the states and their full iterative averages: state convergence alone does not imply that the average enters the comparison chart.  \Cref{lem:averaging-basin} gives a sufficient route for the latter conclusion when the averaging iteration has an invariant contracting basin.  For each path, let $\tau$ denote the last index at which its safeguard is active or its local proposal formula is unavailable; $\tau$ need not be a stopping time.  When $\tau<\infty$ almost surely, the finitely many exact coordinate discrepancies have an almost surely finite sum and therefore contribute $o_p(1)$ after the $n^{-1/2}$ normalization in \eqref{eq:drift-rem-abstract}.  The same statement holds under the joint law for a weighted path and hence conditionally in probability.  Because every weighted trajectory uses its own path-specific finite-exception index and its full iterative average is retained, the reported trajectories remain conditionally independent and identically distributed.  No maximum of entry times over $B_n$ paths is required for empirical inference from the completed statistic in \eqref{eq:Tstar-completed}: for each $n$, the multiplier paths form an i.i.d. row from its conditional law.  In contrast, an implementation that aborts a whole row whenever any inverse is unavailable would require separate rate control and is not used here.  \Cref{lem:finite-prefix-average} records the corresponding invariance of the intrinsic average.

\paragraph{Unweighted expansion.}\label{rem:base-benchmark}
Under smooth RSGD conditions, the unweighted representation is a Polyak-Ruppert benchmark \citep{TripuraneniEtAl2018}.  It is retained because the weighted result must be proved on the same probability space and under the abstract averaged conditions above.  The full-path extension is nontrivial: the weighted oracle is evaluated at a weighted state, yet only the target innovation is multiplied in the first-order representation.

\paragraph{Geometric centering.}\label{rem:geometric-centering}
Equation \eqref{eq:bootstrap-linear-new} is not obtained by formally subtracting manifold-valued points.  The proof first expands both averages in $T_{x_\star}\M$ and then shows that the implemented vector $R_{\bar x_n}^{-1}(\bar x_n^\star)$ is first-order equivalent to their coordinate difference.  This is the geometric step absent in Euclidean COnB.

\begin{proposition}[Exact Euclidean reduction to COnB]\label{prop:euclidean-reduction}
Let $\M=\R^d$, $R_x(v)=x+v$, use no state safeguard and the arithmetic Polyak-Ruppert average, let $\Psi_k(x;D_k)=\nabla h(x,Z_k)$ with i.i.d. data, and take independent $\operatorname{Exp}(1)$ multipliers.  Up to an immaterial indexing convention, \Cref{alg:fullpath} and, for the coordinate functional $\phi_i(x)=x_i$, the interval \eqref{eq:fixedB-CI} are exactly Algorithm~2 and the ASGD confidence interval of \citet{LamWang2026}.  Consequently, \Cref{thm:fixedB,cor:fixedB} recover their fixed-$B$ online coordinatewise result whenever both sets of assumptions hold.
\end{proposition}

\paragraph{Theoretical comparison.}
The overlap in \Cref{prop:euclidean-reduction} is exact at the algorithmic level, while the assumption regimes are nonnested.  \citet{LamWang2026} assume a twice continuously differentiable sample loss, global samplewise strong convexity and bounded Hessian, a sample Hessian that is Lipschitz on a bounded target neighborhood uniformly over observations, a fourth moment for the target gradient, and Glivenko--Cantelli conditions; they allow $\alpha=1$, establish an offline cheap bootstrap and last-iterate results, derive quantitative normal-approximation bounds, and study a sparse high-dimensional extension.  The present theorem is local and fixed-dimensional for averaged iterations with $\alpha<1$, and extends the construction to nonsymmetric derivatives, martingale innovations, samplewise nonsmooth fields controlled in $L^2$, finite-second-moment multipliers, and manifold-valued parameters.

\subsubsection{Large-replication inference}\label{sec:largeB-theory}

When the goal is to estimate the conditional distribution itself, the number of paths must grow.  Weak convergence is enough for continuous quantiles; estimating a covariance asks for the additional tail control stated after the next corollary.

\begin{corollary}[Smooth functionals and intrinsic norm regions]\label{cor:functionals}
Let $\phi:\M\to\R^p$ be differentiable at $x_\star$.  Under \Cref{thm:bootstrap}, conditionally in probability,
\begin{equation}\label{eq:functional-limit}
  \sqrt n\{\phi(\bar x_n^\star)-\phi(\bar x_n)\}
  \rightsquigarrow
  N_p\{0,D\phi_{x_\star}VD\phi_{x_\star}^{\dagg}\},
\end{equation}
and its conditional law consistently estimates the sampling law of
$\sqrt n\{\phi(\bar x_n)-\phi(x_\star)\}$.
If the limiting scalar variance is positive and $B=B_n\to\infty$, \eqref{eq:basic-CI} has asymptotic coverage $1-\tau$.  If the distribution of $\norm Z$, $Z\sim N(0,V)$, is continuous at its $(1-\tau)$ quantile, then \eqref{eq:intrinsic-region} has asymptotic coverage $1-\tau$.
\end{corollary}

\begin{definition}[Conditional uniform integrability in probability]\label{def:cui}
A sequence $(Y_n^\star)$ is conditionally uniformly integrable in probability if, for every $\eps>0$,
\begin{equation}\label{eq:cui-def}
  \lim_{K\to\infty}\limsup_{n\to\infty}
  \Prob\left\{
  \Exp^\star\bigl[|Y_n^\star|\ind\{|Y_n^\star|>K\}\bigr]>\eps
  \right\}=0.
\end{equation}
\end{definition}

\begin{proposition}[Bootstrap covariance and Wald regions]\label{prop:covariance}
Suppose \Cref{thm:bootstrap} holds, $B=B_n\to\infty$, and
$\norm{T_n^\star}^2$ is conditionally uniformly integrable in probability.  Then
\begin{equation}\label{eq:cov-consistency}
  \mathcal P_{\bar x_n\to x_\star}
  \widehat V_{n,B_n}^\star
  \mathcal P_{x_\star\to\bar x_n}
  \xrightarrow{p}V.
\end{equation}
If $V$ is positive definite, the Wald region \eqref{eq:wald-region} has asymptotic coverage $1-\tau$.
\end{proposition}

\paragraph{Additional moment conditions for covariance estimation.}\label{rem:cov-moment}
Conditional uniform integrability of $\norm{T_n^\star}^2$ is needed only to pass from weak convergence to covariance convergence; it is not implied by \eqref{eq:weight-second} and the weak bootstrap theorem.  A transparent sufficient condition is that, for some $\delta>0$,
$\Exp^\star\norm{T_n^\star}^{2+\delta}=O_p(1)$.
Conditional Markov inequality then gives \eqref{eq:cui-def}.  This bound is an additional sufficient condition, not a consequence of the second-moment state bounds in \Cref{prop:state-rate}.  A fourth moment is not required by the general covariance proposition, although a particular quadratic oracle, such as streaming PCA, may itself require a fourth data moment to have a finite innovation variance.

\subsubsection{Primitive conditions for Riemannian stochastic fields}\label{sec:primitive}

The abstract iteration is useful only if its remainders can be verified from an oracle and a manifold update.  We now give one such route.  It replaces samplewise derivatives by a centered local $L^2$ modulus, isolates population smoothness in a deterministic remainder, and uses a separate stability argument for the state rate.

For $u$ in a ball $B_{T_{x_\star}\M}(0,r)$, define the transported oracle
\begin{equation*}
  \psi_k(u)=\mathcal P_{x(u)\to x_\star}\Psi_k\{x(u);D_k\}\in T_{x_\star}\M.
\end{equation*}

\begin{assumption}[Primitive oracle expansion]\label{ass:primitive-oracle}
For $u\in B_{T_{x_\star}\M}(0,r)$,
\begin{equation}\label{eq:primitive-expansion}
  \psi_k(u)=Hu+\xi_k+b_k+\chi_k(u)+r_{\mathrm{pop}}(u),
\end{equation}
where $H$ is nonsingular, $(\xi_k)$ satisfies \Cref{ass:innovation}, $b_k$ is $\F_{k-1}$-measurable and satisfies \eqref{eq:bias-abstract}, and the following conditions hold.
\begin{enumerate}[label=(\roman*),leftmargin=2.2em]
\item $\chi_k(0)=0$.  For every $\G_{k-1}$-measurable $U_{k-1}$ taking values in $B_{T_{x_\star}\M}(0,r)$,
\begin{align}
  \Exp\{\chi_k(U_{k-1})\mid\G_{k-1}\}&=0,
  \label{eq:chi-centered}\\
  \Exp\{\norm{\chi_k(U_{k-1})}^2\mid\G_{k-1}\}
  &\le \omega(\norm{U_{k-1}}),
  \label{eq:chi-L2}
\end{align}
where $\omega$ is bounded on $[0,r]$ and $\omega(t)\to0$ as $t\downarrow0$.  The analogous statements hold for $\F_{k-1}$-measurable arguments.
\item For some $\rho\in(0,1]$,
\begin{equation}\label{eq:r0-holder}
  \norm{r_{\mathrm{pop}}(u)}\le C\norm u^{1+\rho}.
\end{equation}
\item The local oracle has a uniformly bounded conditional second moment:
\begin{equation}\label{eq:oracle-second}
  \sup_{k\ge1}\sup_{u\in B_{T_{x_\star}\M}(0,r)}
  \Exp\{\norm{\psi_k(u)}^2\mid\F_{k-1}\}\le C
  \quad\text{almost surely}.
\end{equation}
The current oracle input $D_k$ is conditionally fresh given the past, so \eqref{eq:chi-centered}--\eqref{eq:oracle-second} remain valid after enlargement by past multipliers.
\end{enumerate}
\end{assumption}

\begingroup The conditional bounds are understood through common measurable conditional-kernel versions. At $u=0$, \eqref{eq:primitive-expansion} and \eqref{eq:oracle-second} give $\Exp(\norm{\xi_k}^2\mid\F_{k-1})+\norm{b_k}^2\le C$, since $\psi_k(0)=\xi_k+b_k$ and $\xi_k$ is centered. Independence of the weight sequence preserves this bound after enlargement to $\G_{k-1}$. Together with \eqref{eq:chi-L2} and the bounded deterministic terms on the local ball, it follows that $\Exp\{\norm{\psi_k(U_{k-1})}^2\mid\G_{k-1}\}\le C'$ for every predictable local argument $U_{k-1}$. Thus the second-moment bound applies at the actual local states without an additional fourth-moment assumption.\endgroup

For i.i.d. data, let $\psi(u;D_k)$ be a measurable random field and $G(u)=\Exp\psi(u;D_k)$.  Then
\begin{equation*}
  \xi_k=\psi(0;D_k)-G(0),
  \qquad
  \chi_k(u)=\psi(u;D_k)-\psi(0;D_k)-\{G(u)-G(0)\}
\end{equation*}
provides the centered decomposition.  The condition on $\chi_k$ is then an ordinary local $L^2$ modulus and does not require a samplewise derivative.

\begin{assumption}[Second-order update compatibility]\label{ass:geometry}
For $u,z\in T_{x_\star}\M$ sufficiently small, define
\begin{equation}\label{eq:Phi-def}
  \Phi(u,z)
  =R_{x_\star}^{-1}
  \left[R_{x(u)}\{\mathcal P_{x_\star\to x(u)}z\}\right].
\end{equation}
There is a constant $C$ such that
\begin{equation}\label{eq:Phi-bound}
  \norm{\Phi(u,z)-u-z}
  \le C\{\norm u^2\norm z+\norm z^2\}.
\end{equation}
\end{assumption}

The exponential map and parallel transport satisfy \Cref{ass:intrinsic-comparison,ass:geometry} by normal-coordinate and Jacobi-field expansions.  A general first-order retraction does not automatically give all three inequalities.  For non-exponential retractions, the chosen retraction, inverse, and transport must be checked together; projection-like second-order retractions on matrix manifolds provide an important class \citep{AbsilMalick2012}.

\begin{proposition}[Primitive verification of the averaged iteration]\label{prop:primitive-verification}
Suppose \Cref{ass:primitive-oracle,ass:intrinsic-comparison,ass:geometry} and the state bounds \eqref{eq:state-abstract} hold, with eventual localization in a common ball where all three assumptions apply.  If a safeguard is used, assume it is eventually inactive for each fixed path; \Cref{prop:state-rate} supplies this conclusion.  If
\begin{equation}\label{eq:alpha-rho}
  \alpha(1+\rho)>1,
\end{equation}
then the actual updates \eqref{eq:base-update} and \eqref{eq:weighted-update} satisfy \Cref{ass:abstract-recursion}, with their finitely many exceptional coordinate discrepancies absorbed into $e_k$ and $e_k^\star$.  Consequently, \Cref{thm:joint,thm:bootstrap,thm:fixedB} hold for the reported full iterative averages.
\end{proposition}

\paragraph{The smoothness threshold.}\label{rem:threshold}
The condition \eqref{eq:alpha-rho} controls
\[
  \frac1{\sqrt n}\sum_{k=1}^n\Exp\norm{u_{k-1}}^{1+\rho}
  \lesssim
  \frac1{\sqrt n}\sum_{k=1}^n\eta_k^{(1+\rho)/2}.
\]
For a locally $C^2$ population field, $\rho=1$, so no restriction is added beyond $\alpha>1/2$.  For $\rho<1$, \eqref{eq:alpha-rho} is sufficient rather than necessary; a direct accumulated remainder bound can replace it.

\begin{proposition}[A second-moment route to the $O(\eta_k)$ state bounds]\label{prop:state-rate}
Let $K\subset\operatorname{int}(K^+)$ be compact neighborhoods of $x_\star$, with $K$ geodesically convex and squared distance twice continuously differentiable on $K^+$. Initialize $x_{0}\in K$ and all weighted paths at the same point.  Suppose the update uses, when necessary, a measurable safeguard $S(x,v)\in K$ for an attempted tangent step $v\in T_x\M$.  There is an $r_0>0$ such that, uniformly for $x\in K$, $R_x(-v)$ is defined and lies in $K^+$ whenever $\norm v_x\le r_0$, and
\[
  d\{S(x,v),x_\star\}\le d\{R_x(-v),x_\star\}
  \quad\text{on short proposals}.
\]
For arbitrary $v$, the safeguard may include tangent-step clipping before applying the retraction, but membership of its output in $K$ is a separate requirement: clipping alone does not imply this invariance.  There are an open neighborhood $U_0$ of $x_\star$, with $\overline U_0\subset K$, and a constant $\delta_0>0$ such that, uniformly for $x\in U_0$,
\begin{equation}\label{eq:safeguard-identity}
  S(x,v)=R_x(-v)\qquad\text{whenever }\norm v_x\le\delta_0.
\end{equation}

Let $G$ be a deterministic mean field. The following conditional bounds use common measurable versions and hold at every predictable $K$-valued state under the original and enlarged filtrations. For some $\mu>0$, suppose uniformly for $x\in K$ that
\begin{align}
  \ip{G(x)}{\Log_x(x_\star)}_x
  &\le-\mu d(x,x_\star)^2,
  \label{eq:strong-monotonicity}\\
  \norm{\Exp\{\Psi_k(x;D_k)\mid\F_{k-1}\}-G(x)}_x
  &\le a_k,
  \qquad a_k^2\le C\eta_k,
  \label{eq:mean-bias-state}\\
  \Exp\{\norm{\Psi_k(x;D_k)}_x^2\mid\F_{k-1}\}
  &\le C.
  \label{eq:oracle-second-state}
\end{align}
Assume the current oracle is conditionally fresh given the past, and the multipliers satisfy \eqref{eq:weight-second}.  For sufficiently small $\eta_0$, the localized paths satisfy
\begin{equation}\label{eq:state-rate-conclusion}
  \Exp d(x_k,x_\star)^2\le C\eta_k,
  \qquad
  \Exp\Exp^\star d(x_k^\star,x_\star)^2\le C\eta_k.
\end{equation}
Moreover, the safeguard is applied only finitely often almost surely for each fixed path.
\end{proposition}

The preceding proposition controls the states.  The next deterministic result supplies the missing implication for the unchanged full average whenever the chosen geometry admits an invariant averaging basin.

\begin{lemma}[Averaging-basin localization of the full average]\label{lem:averaging-basin}
Let $\alpha\in(0,1)$ and let $K\subset\M$ contain $x_\star$.  Suppose that, for every $x,y\in K$ and $t\in[0,1]$, the inverse retraction is defined, the averaging map
$\mathsf A_t(x,y)=R_x\{tR_x^{-1}(y)\}$
belongs to $K$, and, for constants $\kappa\in(\alpha,1]$ and $C<\infty$,
\begin{equation}\label{eq:averaging-basin-contraction}
  d\{\mathsf A_t(x,y),x_\star\}^2
  \le (1-\kappa t)d(x,x_\star)^2+Ct\,d(y,x_\star)^2.
\end{equation}
If a state path $(x_k)$ lies in $K$, converges to $x_\star$, and its full iterative average satisfies $\bar x_1=x_1$ and $\bar x_k=\mathsf A_{1/k}(\bar x_{k-1},x_k)$, then $\bar x_k$ remains in $K$ and converges to $x_\star$.  If additionally $\Exp d(x_k,x_\star)^2\le C_0k^{-\alpha}$, then $\Exp d(\bar x_k,x_\star)^2\le C_1k^{-\alpha}$.  The same conclusions hold under the joint original--multiplier law after replacing $\Exp$ by $\Exp\Exp^\star$.
\end{lemma}

This lemma is a sufficient primitive route to the average-localization clause of \Cref{ass:intrinsic-comparison}; the abstract theorem continues to allow that clause to be verified directly.

\begingroup\paragraph{A concrete safeguard and averaging basin.}\label{rem:hadamard-safeguard}
On a Hadamard manifold, choose a specified compact geodesically convex set $K$ containing $x_\star$ in its interior and a fixed $c\in K$. For a short proposal use $S(x,v)=\Pi_K\{R_x(-v)\}$, where $\Pi_K$ is metric projection; for a long or unavailable proposal set $S(x,v)=c$. Projection does not increase distance to any point of $K$, so the output is invariant in $K$ and the distance inequality in \Cref{prop:state-rate} holds. Compactness supplies a uniform short-step neighborhood, and sufficiently small steps from a neighborhood of $x_\star$ already lie in $K$, giving \eqref{eq:safeguard-identity}. The rule evaluates neither $x_\star$ nor an entry time; $x_\star\in\operatorname{int}K$ is a model restriction. After verifying the drift and moment bounds on $K$, \Cref{prop:state-rate} yields convergence and eventual inactivity. For exponential averaging, geodesic convexity of squared distance gives \eqref{eq:averaging-basin-contraction} with $\kappa=C=1$. This completes the state and full-average verification without assuming either conclusion.\endgroup

\paragraph{Why stability is stronger than nonsingularity.}\label{rem:strong-mono}
Strong monotonicity is not part of the abstract bootstrap theorem.  It is one route to the state rate.  It follows locally from a positive-definite Riemannian Hessian for geodesically convex objectives and from an eigengap for principal-subspace estimation.  A model-specific Lyapunov function or contraction argument may replace it.  The bound $a_k^2\lesssim\eta_k$ is used only for state stability; inference still requires the accumulated-bias condition \eqref{eq:bias-abstract}.

\subsubsection{How the assumptions divide the work}\label{sec:assumption-discussion}

The abstract theorem takes the state rate and complete-path localization as inputs. A primitive route first verifies the drift and moments for an actual invariant safeguard, then obtains state convergence from \Cref{prop:state-rate} and full-average localization from \Cref{lem:averaging-basin}; only then does \Cref{prop:primitive-verification} supply the averaged iteration. The following distinctions explain which assumptions serve inference and which provide sufficient verification devices.  Nonsingularity of $H$ is structural: without a locally invertible derivative, the usual linear influence representation need not hold.  Symmetry and positive definiteness are not required by the abstract Abel argument.  They enter only through a chosen stability proof, as in strongly geodesically convex objectives or eigengap problems.  This distinction is standard in stochastic approximation, where the averaging limit and the Lyapunov analysis need not use identical hypotheses \citep{PolyakJuditsky1992,TripuraneniEtAl2018}.

The step range $\alpha\in(1/2,1)$ is conventional for averaged stochastic approximation.  The lower bound makes $\sum\eta_k^2<\infty$ and controls geometric one-step errors; the upper bound removes the Abel boundary and is used in the $O(\eta_k)$ comparison induction.  The endpoint $\alpha=1$ is possible in Euclidean and Riemannian models, but the admissible step constant then depends on the spectrum of $H$ and requires a separate argument.  \citet{LamWang2026}, for example, include $\alpha=1$ under their global smooth strongly convex assumptions.  More general regularly varying steps can be used whenever the summation-by-parts bounds in \Cref{lem:abel-boundary} remain valid.

The martingale covariance and Lindeberg conditions are close to the standard minimal assumptions for a finite-dimensional martingale central limit theorem \citep{HallHeyde1980}.  A $2+\delta$ score moment is a transparent sufficient condition for the Lindeberg condition, not a required high-order assumption.  The same applies to the multipliers: mean one preserves the population drift, variance one calibrates $Q$, and a finite second moment is enough for the conditional Lindeberg argument.  The fourth moment in the Grassmann application is different: there the stochastic gradient is quadratic in the data, so a fourth data moment supplies a sufficient square-integrable envelope for the score and its local increments.

The primitive $C^{1,\rho}$ condition is imposed on the population field, not on every sample loss.  It can be replaced by the direct accumulated remainder in \eqref{eq:drift-rem-abstract}.  The local $L^2$ modulus similarly permits nonsmooth sample paths; the geometric median application below is an example.  Relative to the quantitative Euclidean analysis of \citet{LamWang2026}, these conditions replace global samplewise Hessian control by local population and $L^2$ requirements, while adding explicit state, localization, and geometric hypotheses.  The assumption sets are therefore non-nested, and the present conclusion is only first-order and fixed-dimensional.

The geometric inequalities \eqref{eq:A-bound}, \eqref{eq:Gamma-bound}, and \eqref{eq:Phi-bound} are local second-order coordinate identities.  They are automatic for exponential-map normal coordinates with parallel transport and can be checked for common second-order matrix retractions \citep{AbsilEtAl2008,AbsilMalick2012}.  They are not probabilistic smoothness assumptions.  Localization is a substantive path condition, separate from these deterministic identities; \Cref{prop:state-rate} shows eventual safeguard inactivity separately for each path, and \Cref{lem:finite-prefix-average} covers the unchanged full averages without an online or common restart time.

Finally, the bias conditions are substantive.  Pointwise convergence $b_k\to0$ is insufficient because the Polyak-Ruppert expansion accumulates $b_k$ before dividing by $\sqrt n$.  \Cref{prop:bias} proves the fixed boundary statement under the abstract state conditions and records the subcritical Abel term only as a coordinate-level diagnostic pending a bias-aware manifold remainder analysis.  Conditional uniform integrability is needed only for covariance estimation, not for fixed-$B$ pivotal intervals or large-$B$ quantile inference.

\paragraph{Bias accumulates before averaging.}\label{rem:bias-diagnostic}
The multiplier perturbation of the bias has conditional variance $n^{-1}\sum\norm{b_k}^2$, which vanishes for every polynomial rate $k^{-\gamma}$ with $\gamma>0$.  It therefore cannot mimic a fixed or diverging deterministic shift in the sampling law.

\begingroup Let $h_k=h_0k^{-\beta}$, with $h_0,r,\beta>0$ as in \Cref{prop:bias}. For a state-dependent approximation bias $h_k^rB(u)$, controlling only $b_k=h_k^rB(0)$ leaves an additional drift $h_k^r\{B(u_{k-1})-B(0)\}$. If $B$ is locally Lipschitz and the state rate holds, its root-$n$ accumulated expectation is bounded by $Cn^{-1/2}\sum_{k\le n}h_k^r\eta_k^{1/2}$, which vanishes when $\alpha+2r\beta>1$. This is a sufficient check on the remaining drift, not a replacement for bias-aware state stability in a subcritical regime.\endgroup

\subsection{Non-Euclidean Applications}\label{sec:applications}

The examples below use i.i.d. observations and exponential-map notation for clarity.  A non-exponential implementation may be substituted only after its retraction, local inverse, and transport jointly satisfy the three geometric comparisons.  \Cref{lem:implementation-geometry} verifies the normalized spherical projection and Grassmann QR/graph-inverse triples used below.  In each case the displayed limiting law is also the conditional limit of the full-path multiplier difference, and both the fixed-$B$ and large-$B$ procedures apply under their respective moment and nondegeneracy conditions.

\subsubsection{Intrinsic means on a sphere}\label{sec:app-sphere}

Let $\M=\mathbb S^q$ with its canonical metric and
$F(x)=\frac12\Exp d(x,Y)^2$.
Assume that, for some $x_\star\in\mathbb S^q$ and $r_0<\pi/2$,
\begin{equation}\label{eq:sphere-support}
  \Prob\{d(x_\star,Y)\le r_0\}=1,
  \qquad \Exp\Log_{x_\star}(Y)=0.
\end{equation}
Let $v=\Log_{x_\star}(Y)$, $r=\norm v$, and $e=v/r$ when $r>0$.  With the entire Hessian integrand defined to be $I$ at $r=0$, define
\begin{align}
  H_{\mathbb S}
  &=\Exp\left[e\ot e+r\cot(r)\{I-e\ot e\}\right],
  \label{eq:sphere-H}\\
  Q_{\mathbb S}&=\Exp(v\ot v).
  \label{eq:sphere-Q}
\end{align}
The support condition excludes the cut locus and makes every eigenvalue in the integrand of \eqref{eq:sphere-H} positive.  It is a transparent sufficient condition for a regular local mean, not the most general condition for a spherical Fr\'echet-mean central limit theorem; cut-locus configurations can require additional terms \citep{HotzLeWood2024}.

\begin{corollary}[Spherical intrinsic mean]\label{cor:sphere}
Suppose \eqref{eq:sphere-support} holds, the original and weighted states and full iterative averages satisfy the localization clause of \Cref{ass:intrinsic-comparison} in a common strongly convex neighborhood of $x_\star$, and their state rates satisfy \eqref{eq:state-abstract}.  Use the stochastic gradient
$\Psi(x;Y)=-\Log_x(Y)$.
Then
\begin{equation}\label{eq:sphere-limit}
  \sqrt n\,\Log_{x_\star}(\bar x_n)
  \rightsquigarrow
  N\left(0,H_{\mathbb S}^{-1}Q_{\mathbb S}H_{\mathbb S}^{-1}\right),
\end{equation}
and the multiplier law is conditionally consistent.  The fixed-$B$ scalar intervals in \Cref{cor:fixedB} are valid for differentiable spherical functionals with nonzero limiting variance; the full intrinsic region additionally requires the limiting covariance in \eqref{eq:sphere-limit} to be positive definite.
\end{corollary}

The localization condition in \Cref{cor:sphere} has a direct sufficient verification for the two spherical averaging maps used most often in practice.

\begin{corollary}[A cap-preserving route for spherical full averages]\label{cor:sphere-averaging-basin}
Let $K_\rho=\{x\in\mathbb S^q:d(x,x_\star)\le\rho\}$ with $\rho<\pi/4$.  Suppose the original state and every fixed multiplier state are initialized in $K_\rho$, are kept in $K_\rho$ by a measurable cap-preserving localization rule, converge almost surely to $x_\star$, and satisfy the $O(k^{-\alpha})$ second-moment state bounds under the original and joint laws.  If the full iterative averages use either the exponential map and logarithm or the normalized projection retraction and its local inverse, then every full average remains in $K_\rho$, converges almost surely to $x_\star$, and has the $O(k^{-\alpha})$ second-moment rate.  Consequently, each base or fixed multiplier average enters the comparison chart after its own almost surely finite index, without a restart or a maximum entry time across multiplier paths.
\end{corollary}

The cap-membership clause concerns the output of the localization rule.  Rescaling a tangent proposal controls its norm but does not by itself certify that the retracted state remains in $K_\rho$.

The radius $r_0<\pi/2$ is stronger than uniqueness alone, but it gives cut-locus avoidance, a smooth bounded score, and positive local curvature in one condition; related convexity radii are standard in the theory of Riemannian centers of mass \citep{Afsari2011}.  If the support approaches the antipode, neither the displayed Hessian nor a standard Gaussian limit should be used without a separate cut-locus analysis.

\subsubsection{Hadamard Fr\'echet means and positive-definite matrices}\label{sec:app-hadamard}

Let $\M$ be a finite-dimensional Hadamard manifold and let
\begin{equation*}
  F(x)=\frac12\Exp d(x,Y)^2,
  \qquad x_\star=\operatorname*{arg\,min}_xF(x).
\end{equation*}
Assume that $Y$ is supported in a compact set, and localize the iterations to a compact geodesically convex neighborhood containing $x_\star$.  Hadamard geometry removes the cut locus and makes squared distance geodesically strongly convex \citep{Bacak2014}.  Put
\begin{equation*}
  H_F=\Hess F(x_\star),
  \qquad
  Q_F=\Exp\{\Log_{x_\star}(Y)\ot\Log_{x_\star}(Y)\}.
\end{equation*}

\begin{corollary}[Hadamard Fr\'echet mean]\label{cor:hadamard}
Under the preceding support and compact-localization conditions, suppose the original and weighted states and full iterative averages satisfy the localization clause of \Cref{ass:intrinsic-comparison}.  Use
$\Psi(x;Y)=-\Log_x(Y)$ and suppose the $O(\eta_k)$ state rates hold.  Then
\begin{equation}\label{eq:hadamard-limit}
  \sqrt n\,\Log_{x_\star}(\bar x_n)
  \rightsquigarrow
  N(0,H_F^{-1}Q_FH_F^{-1}),
\end{equation}
and the full-path multiplier law is conditionally consistent.
\end{corollary}

A principal instance is the $d=m(m+1)/2$ dimensional manifold $\SPD_m$ of $m\times m$ symmetric positive-definite matrices with the affine-invariant metric $\ip{\mathrm{U}}{\mathrm{V}}_{\mathrm{X}}=\tr(\mathrm{X}^{-1}\mathrm{U}\mathrm{X}^{-1}\mathrm{V})$. Its exponential and logarithm maps are
\begin{align}
  \Expmap_{\mathrm{X}}(\mathrm{U})
  &=\mathrm{X}^{1/2}\exp(\mathrm{X}^{-1/2}\mathrm{U}\mathrm{X}^{-1/2})\mathrm{X}^{1/2},
  \label{eq:spd-exp}\\
  \Log_{\mathrm{X}}(\mathrm{Y})
  &=\mathrm{X}^{1/2}\log(\mathrm{X}^{-1/2}\mathrm{Y}\mathrm{X}^{-1/2})\mathrm{X}^{1/2}.
  \label{eq:spd-log}
\end{align}
This geometry is complete, simply connected, and nonpositively curved \citep{PennecFillardAyache2006}.  If the eigenvalues of $\mathrm{Y}$ lie almost surely in a fixed interval $[a,b]\subset(0,\infty)$, all local score and smoothness conditions hold on an eigenvalue-bounded neighborhood.  For $\phi(\mathrm{X})=\log\det \mathrm{X}$,
\begin{equation}\label{eq:logdet-derivative}
  D\phi_{\mathrm{X}_\star}[\mathrm{U}]=\tr(\mathrm{X}_\star^{-1}\mathrm{U}),
\end{equation}
so \Cref{cor:fixedB,cor:functionals} give direct intervals for the log generalized volume without an explicit estimator of $H_F$, provided $D\phi_{\mathrm{X}_\star}V(D\phi_{\mathrm{X}_\star})^\dagg>0$.
For $\phi_{\max}(\mathrm{X})=\log\lambda_{\max}(\mathrm{X})$, the same scalar theory applies only when the largest eigenvalue of $\mathrm{X}_\star$ is simple.  If $v_{\max}$ is its normalized leading eigenvector and the limiting variance is nonzero, then
\begin{equation}\label{eq:logmax-derivative}
  D\phi_{\max,\mathrm{X}_\star}[\mathrm{U}]
  =\frac{v_{\max}^{\trans}\mathrm{U}v_{\max}}{\lambda_{\max}(\mathrm{X}_\star)}.
\end{equation}
A positive target eigengap is therefore an explicit regularity condition for the largest-log-eigenvalue intervals below; no such inference is claimed at a repeated top eigenvalue.

\begin{corollary}[Affine-invariant log-normal moment route]\label{cor:spd-lognormal}
Let $\mathrm{E}$ be a symmetric random matrix whose law is non-atomic, with $\mathrm{E}\stackrel{d}{=}-\mathrm{E}$ and $\Exp\norm{\mathrm{E}}_F^{2+\delta}<\infty$ for some $\delta>0$, and set
$\mathrm{Y}=\mathrm{A}^{1/2}\exp(\mathrm{E})\mathrm{A}^{1/2},\qquad \mathrm{A}\in\SPD_m$.
If the original and weighted states and full iterative averages are localized in a common compact neighborhood of $\mathrm{A}$ in the sense of \Cref{ass:intrinsic-comparison}, satisfy \eqref{eq:state-abstract}, and the population mean field satisfies \eqref{eq:r0-holder} for some $\rho\in(0,1]$ with $\alpha(1+\rho)>1$ and nonsingular derivative, then $\mathrm{A}$ is the unique affine-invariant Fr\'echet mean and the conclusions of \Cref{cor:hadamard} hold.  For the normalized median score, assume separately that its original and weighted states and full iterative averages satisfy the same localization requirements, that its states satisfy \eqref{eq:state-abstract}, that $\mathrm{A}$ is the unique median, and that its population field is locally $C^{1,\rho}$ with nonsingular derivative and $\alpha(1+\rho)>1$. Then \Cref{cor:median} applies to the median iteration.
\end{corollary}

The bounded-eigenvalue condition packages compact localization, finite score moments, and uniform derivative bounds, but is not necessary.  \Cref{cor:spd-lognormal} gives an unbounded score-moment route: on a compact state set, the affine-invariant score and its local derivative admit an envelope of order $1+\norm{\mathrm{E}}_F$, so the stated moment controls the score second moment and local $L^2$ modulus.  The population remainder \eqref{eq:r0-holder} is still required separately.  The unweighted Gaussian limit for intrinsic means is classical \citep{BhattacharyaPatrangenaru2003,BhattacharyaPatrangenaru2005}; the result here supplies the online full-path conditional law and fixed-replication calibration.

\subsubsection{Riemannian geometric medians}\label{sec:app-median}

Let $\M$ be a finite-dimensional Hadamard manifold, assume $\Exp d(x_{0},Y)<\infty$ for some $x_{0}\in\M$, and consider the population objective $x\mapsto\Exp d(x,Y)$.
Suppose $x_\star$ is its unique minimizer, $\Prob(Y=x_\star)=0$, and the population field
\begin{equation*}
  G_{\mathrm{med}}(x)
  =-\Exp\left\{\frac{\Log_x(Y)}{d(x,Y)}\right\}
\end{equation*}
exists and is $C^{1,\rho}$ in a neighborhood of $x_\star$, with nonsingular derivative
$H_{\mathrm{med}}=DG_{\mathrm{med}}(x_\star)$.  Define
\begin{equation*}
  Q_{\mathrm{med}}
  =\Exp\left[
  \frac{\Log_{x_\star}(Y)}{d(x_\star,Y)}
  \ot
  \frac{\Log_{x_\star}(Y)}{d(x_\star,Y)}
  \right].
\end{equation*}
The stochastic score has norm one away from $Y=x$ and is samplewise nonsmooth at the observation.

\begin{corollary}[Riemannian geometric median]\label{cor:median}
Use the score
\begin{equation*}
  \Psi(x;Y)=-\frac{\Log_x(Y)}{d(x,Y)},
  \qquad Y\ne x,
\end{equation*}
with any measurable convention of uniformly bounded norm at $Y=x$.  Suppose the original and weighted states and full iterative averages satisfy the localization clause of \Cref{ass:intrinsic-comparison}, the states satisfy the local rates in \eqref{eq:state-abstract}, $\alpha(1+\rho)>1$, and the law of $Y$ is non-atomic on their common localized neighborhood.  Thus, for every fixed path and time, the exceptional events $Y_k=x_{k-1}$ and $Y_k=x_{k-1}^\star$ have probability zero.  Then
\begin{equation}\label{eq:median-limit}
  \sqrt n\,\Log_{x_\star}(\bar x_n)
  \rightsquigarrow
  N\left(0,H_{\mathrm{med}}^{-1}Q_{\mathrm{med}}H_{\mathrm{med}}^{-\dagg}\right),
\end{equation}
and the conditional multiplier approximation is valid.  Fixed-$B$ scalar intervals and full intrinsic regions follow under, respectively, the nonzero-variance and positive-definiteness conditions of \Cref{cor:fixedB}.
\end{corollary}

Existence, uniqueness, and algorithms for Riemannian medians are studied by \citet{FletcherEtAl2009} and \citet{Yang2010}.  The substantive condition here is differentiability of the \emph{population} field, not of the sample distance.  A concrete sufficient model, including the population remainder and state stability, is given in \Cref{prop:median-model}.  Non-atomicity excludes equality at either predictable state, and dominated convergence yields the local $L^2$ modulus.  The example therefore shows how first-order full-path inference can accommodate a samplewise nonsmooth score.

\begingroup\begin{proposition}[A verifiable nonsmooth median model]\label{prop:median-model}
On $\SPD_m$ with the affine-invariant metric, let $m\ge2$, $d=m(m+1)/2$, and fix $\mathrm{A}\in\SPD_m$. Suppose $\mathrm{Y}$ has density $p_{\mathrm{A}}(y)=c_Rh\{d(\mathrm{A},y)^2\}$ with respect to Riemannian volume, where $c_R$ normalizes the density, $R>0$, and
\[
 h(s)=\begin{cases}\exp\{-1/(R^2-s)\},&s<R^2,\\0,&s\ge R^2.\end{cases}
\]
Then $\mathrm{A}$ is its unique geometric median, the population median field is locally $C^{1,1}$, and its derivative and innovation covariance at $\mathrm{A}$ are positive definite. Choose a specified compact geodesically convex $K$ with $\mathrm{A}\in\operatorname{int}K$, initialize in $K$, and use the projection safeguard {described \hyperref[rem:hadamard-safeguard]{above}} with exponential updates and averaging. For sufficiently small $\eta_0$, $\alpha\in(1/2,1)$, and independent multipliers satisfying \eqref{eq:weight-second}, the state-rate, full-average localization, and primitive conditions hold. Hence \Cref{cor:median} and its fixed-$B$ conclusions apply, including with signed multipliers.
\end{proposition}

The proof in \Cref{app:applications} verifies the conditions from a compact smooth density. This example establishes attainability of the nonsmooth theory; the separate unbounded log-normal model in \Cref{cor:spd-lognormal} retains its explicit population-regularity and localization hypotheses.\endgroup

\subsubsection{Streaming principal-subspace estimation}\label{sec:app-grassmann}

Let $Y\in\R^p$ be i.i.d., centered, with covariance
\begin{equation*}
  \mathrm{\Sigma}=[\mathrm{U}_\star,\mathrm{U}_\perp]
  \begin{pmatrix}\mathrm{\Lambda}_1&0\\0&\mathrm{\Lambda}_2\end{pmatrix}
  [\mathrm{U}_\star,\mathrm{U}_\perp]^{\trans},
\end{equation*}
where $\mathrm{U}_\star$ spans the leading $r$-dimensional eigenspace and
$\lambda_r>\lambda_{r+1}$.  On $\Gr(r,p)$, consider
\begin{equation}\label{eq:pca-objective}
  F([\mathrm{U}])=-\frac12\Exp\norm{\mathrm{U}^{\trans}Y}^2,
\end{equation}
with sample Riemannian gradient $\Psi([\mathrm{U}];Y)=-(\mathrm{I}-\mathrm{U}\mathrm{U}^{\trans})YY^{\trans}\mathrm{U}$. A horizontal tangent vector at $[\mathrm{U}_\star]$ is $\mathrm{U}_\perp \mathrm{K}$, with
$\mathrm{K}\in\R^{(p-r)\times r}$.  Define
\begin{align}
  \mathcal H(\mathrm{K})&=\mathrm{K}\mathrm{\Lambda}_1-\mathrm{\Lambda}_2\mathrm{K},
  \label{eq:pca-H}\\
  \mathrm{Q}_{\mathrm{PCA}}
  &=\Cov\{\vecop(Y_2Y_1^{\trans})\},
  \qquad
  Y_1=\mathrm{U}_\star^{\trans}Y,\quad Y_2=\mathrm{U}_\perp^{\trans}Y.
  \label{eq:pca-Q}
\end{align}
The eigengap makes $\mathcal H$ positive definite.  Let $\mathrm{H}_{\mathrm{vec}}$ denote its $r(p-r)\times r(p-r)$ matrix in the vectorized coordinates $\vecop(\mathrm{K})$.

\begin{corollary}[Grassmann principal-subspace bootstrap]\label{cor:grassmann}
Assume $\Exp\norm Y^4<\infty$.  Use the exponential map and parallel transport, or a named retraction--inverse--transport triple verified to satisfy \eqref{eq:A-bound}, \eqref{eq:Gamma-bound}, and \eqref{eq:Phi-bound}.  Suppose the original and weighted states and full iterative averages satisfy the localization clause of \Cref{ass:intrinsic-comparison} in the attraction basin of $[\mathrm{U}_\star]$, and the states satisfy \eqref{eq:state-abstract}.  Then
\begin{equation}\label{eq:pca-limit}
  \sqrt n\,\vecop\{\mathrm{U}_{\perp}^{\trans}R_{[\mathrm{U}_\star]}^{-1}(\bar {\mathrm{U}}_n)\}
  \rightsquigarrow
  N\left(0,\mathrm{H}_{\mathrm{vec}}^{-1}\mathrm{Q}_{\mathrm{PCA}}\mathrm{H}_{\mathrm{vec}}^{-\trans}\right),
\end{equation}
and the multiplier law is conditionally consistent.

If $Y$ is Gaussian in the eigenbasis, tangent coordinate $(a,j)$ has asymptotic variance
\begin{equation}\label{eq:pca-gaussian-variance}
  \frac{\lambda_{r+a}\lambda_j}
       {(\lambda_j-\lambda_{r+a})^2},
  \qquad a=1,\ldots,p-r,\quad j=1,\ldots,r.
\end{equation}
For a symmetric matrix $\mathrm{A}$ and the projector functional
$\phi_{\mathrm{A}}([\mathrm{U}])=\tr(\mathrm{A}\mathrm{U}\mathrm{U}^{\trans})$, the differential at $[\mathrm{U}_\star]$ is
\begin{equation}\label{eq:projection-delta}
  D\phi_{\mathrm{A}}[\mathrm{U}_\perp \mathrm{K}]
  =2\ip{\mathrm{U}_\perp^{\trans}\mathrm{A}\mathrm{U}_\star}{\mathrm{K}}_{F}.
\end{equation}
Thus \Cref{cor:fixedB} gives a fixed-replication interval for a specified projector contrast whenever $D\phi_{\mathrm{A}} V(D\phi_{\mathrm{A}})^\dagg>0$. A contrast with zero differential, such as $\mathrm{A}=\mathrm{I}$, is not covered by the nondegenerate scalar pivot.
\end{corollary}

The fourth data moment supplies a uniform square-integrable envelope for the quadratic score and its local increments; it is sufficient, rather than equivalent to a target-coordinate innovation moment.  The eigengap is the local identifiability condition, and \eqref{eq:pca-gaussian-variance} shows its inverse-square effect on uncertainty.  The scalar delta method applies to smooth projector contrasts; principal-angle maxima are nonregular at coincident subspaces and require a different limit theory.

Existing Riemannian averaging and generalized-eigenvector results provide the unweighted local benchmark \citep{TripuraneniEtAl2018,LiJordan2023}.  Rank-one online Oja bootstrap and entrywise inference are studied by \citet{LundeSarkarWard2021} and \citet{KumarPandeySarkar2025}.  The general-rank online multiplier method of \citet{XuLi2026} concerns the high-dimensional final iterate with a run-constant step and bounded nonnegative weights.  The result above follows from a fixed-dimensional decreasing-step Polyak-Ruppert theorem, permits signed finite-second-moment multipliers for weak validity, and supplies fixed-$B$ pivots for smooth contrasts.

\subsection{Proofs}\label{app:proofs}
The following subsections collect every proof and analytic derivation. Result statements needed within a proof are included at their point of use.

\subsubsection{Geometric and Averaging Lemmas}\label{app:geometry}

This appendix proves the deterministic and probabilistic lemmas used in the main argument.  Throughout, constants may change from line to line.

\begin{lemma}[Abel boundary]\label{lem:abel-boundary}
Let $\eta_k=\eta_0k^{-\alpha}$ with $\alpha\in(0,1)$ and suppose
$\Exp\norm{u_k}^2\le C\eta_k$.  Then
\begin{equation}\label{eq:abel-boundary}
  \frac1{\sqrt n}\sum_{k=1}^n
  \frac{u_{k-1}-u_k}{\eta_k}=o_{L^1}(1).
\end{equation}
The same conclusion holds under the joint original--multiplier law if
$\Exp\Exp^\star\norm{u_k^\star}^2\le C\eta_k$.
\end{lemma}

\begin{proof}
Summation by parts gives
\begin{align*}
  \sum_{k=1}^n\frac{u_{k-1}-u_k}{\eta_k}
  &=\frac{u_0}{\eta_1}
    +\sum_{k=1}^{n-1}u_k\left(\frac1{\eta_{k+1}}-\frac1{\eta_k}\right)
    -\frac{u_n}{\eta_n}.
\end{align*}
By Cauchy--Schwarz, $\Exp\norm{u_k}\le C\eta_k^{1/2}=Ck^{-\alpha/2}$.  Since
$\eta_{k+1}^{-1}-\eta_k^{-1}\le Ck^{\alpha-1}$,
\begin{align*}
  \frac1{\sqrt n}\sum_{k=1}^{n-1}
  \Exp\norm{u_k}
  \left(\frac1{\eta_{k+1}}-\frac1{\eta_k}\right)
  &\le \frac C{\sqrt n}\sum_{k=1}^{n-1}k^{\alpha/2-1}
   \le Cn^{(\alpha-1)/2}\to0.
\end{align*}
The terminal term satisfies
\[
  \frac{\Exp\norm{u_n}}{\sqrt n\eta_n}
  \le Cn^{-1/2}n^{\alpha}n^{-\alpha/2}
  =Cn^{(\alpha-1)/2}\to0,
\]
and the initial term is $O(n^{-1/2})$.  This proves \eqref{eq:abel-boundary}.  Replacing $\Exp$ by $\Exp\Exp^\star$ gives the weighted statement.
\end{proof}

\begin{lemma}[Local expansion of the iterative intrinsic average]\label{lem:average-expansion}
Suppose \eqref{eq:A-bound} holds on $B_{T_{x_\star}\M}(0,r)$.  Let
$v_k=R_{x_\star}^{-1}(\bar x_k)$ and assume $u_k,v_k\in B_{T_{x_\star}\M}(0,r)$.  Then
\begin{equation}\label{eq:average-recursion-coordinate}
  v_k=(1-k^{-1})v_{k-1}+k^{-1}u_k+\Delta_k,
  \qquad
  \norm{\Delta_k}\le \frac Ck(\norm{v_{k-1}}^2+\norm{u_k}^2).
\end{equation}
If $\Exp\norm{u_k}^2\le Ck^{-\alpha}$ for $\alpha\in(0,1)$ and $r$ is sufficiently small, then
\begin{equation}\label{eq:v-rate}
  \Exp\norm{v_k}^2\le Ck^{-\alpha}.
\end{equation}
Consequently, if $\alpha>1/2$,
\begin{equation*}
  \sqrt n\left(v_n-\frac1n\sum_{k=1}^nu_k\right)=o_{L^1}(1).
\end{equation*}
All conclusions hold under the joint law for a weighted trajectory.
\end{lemma}

\begin{proof}
The definition \eqref{eq:base-average} gives $v_k=A_{1/k}(v_{k-1},u_k)$, and \eqref{eq:A-bound} yields \eqref{eq:average-recursion-coordinate}.  Put
$w_k=(1-k^{-1})v_{k-1}+k^{-1}u_k$.  Convexity of the squared norm gives
\begin{equation*}
  \norm{w_k}^2
  \le(1-k^{-1})\norm{v_{k-1}}^2+k^{-1}\norm{u_k}^2.
\end{equation*}
For $u_k,v_{k-1}\in B(0,r)$, \eqref{eq:average-recursion-coordinate} implies
$\norm{\Delta_k}\le Crk^{-1}(\norm{v_{k-1}}+\norm{u_k})$.  Expanding
$\norm{w_k+\Delta_k}^2$ and increasing $C$ gives
\[
  \norm{v_k}^2
  \le\left(1-\frac{1-Cr}{k}\right)\norm{v_{k-1}}^2
     +\frac{1+Cr}{k}\norm{u_k}^2.
\]
Choose $r$ so that $\lambda=1-Cr>\alpha$.  With $A_k=\Exp\norm{v_k}^2$, $A_k\le(1-\lambda/k)A_{k-1}+Ck^{-1-\alpha}$. A standard comparison, obtained by multiplying by $k^\lambda$ and summing, gives $A_k\le Ck^{-\alpha}$, proving \eqref{eq:v-rate}.

Multiplying \eqref{eq:average-recursion-coordinate} by $k$ and telescoping yields
\begin{equation*}
  v_n-\frac1n\sum_{k=1}^nu_k
  =\frac1n\sum_{k=2}^nk\Delta_k.
\end{equation*}
Hence
\begin{align*}
  \Exp\norm{\sqrt n(v_n-n^{-1}\textstyle\sum_{k=1}^nu_k)}
  &\le \frac C{\sqrt n}\sum_{k=2}^n
  \{\Exp\norm{v_{k-1}}^2+\Exp\norm{u_k}^2\}\\
  &\le \frac C{\sqrt n}\sum_{k=1}^nk^{-\alpha}
  =O(n^{1/2-\alpha})\to0.
\end{align*}
The proof is pathwise up to expectations and therefore applies under the joint law.
\end{proof}

\begin{lemma}[Finite-prefix invariance of the intrinsic average]\label{lem:finite-prefix-average}
Let $u_k=R_{x_\star}^{-1}(x_k)$ and $v_k=R_{x_\star}^{-1}(\bar x_k)$ be the coordinates of the actual state and full iterative average.  Suppose the state bound in \eqref{eq:state-abstract} holds and there is an almost surely finite, path-specific index $\tau$ such that $u_k,v_{k-1}\in B_{T_{x_\star}\M}(0,r)$ and \eqref{eq:average-recursion-coordinate} holds for every $k>\tau$. Assume $r$ is sufficiently small that the squared-norm comparison in the proof of \Cref{lem:average-expansion} holds with some $\lambda\in(\alpha,1)$. Then, for $\alpha>1/2$,
\begin{equation}\label{eq:finite-prefix-average}
  \sqrt n\left(v_n-\frac1n\sum_{k=1}^n u_k\right)=o_p(1).
\end{equation}
The same conclusion is $o_{\Prob^\star}(1)$ in probability for a weighted trajectory under the joint state bound.  The result uses the reported full average and does not restart it at $\tau$.
\end{lemma}

\begin{proof}
For $n>\tau$, multiplying \eqref{eq:average-recursion-coordinate} by $k$ and telescoping only over the indices after $\tau$ gives the exact identity
\[
  v_n-\frac1n\sum_{k=1}^n u_k
  =\frac{\tau v_\tau-\sum_{k=1}^\tau u_k}{n}
   +\frac1n\sum_{k=\tau+1}^n k\Delta_k.
\]
The first numerator is an almost surely finite random vector. Take $\tau\ge1$ without loss of generality. For $M\ge1$ and $L<\infty$, put
\[
 E_{M,L}=\{\tau\le M,\ \max_{1\le j\le M}\norm{v_j}\le L\}.
\]
On this event, for every $k>M$, the pathwise squared-norm comparison gives
\[
 \norm{v_k}^2\ind(E_{M,L})
 \le (1-\lambda/k)\norm{v_{k-1}}^2\ind(E_{M,L})
       +(C/k)\norm{u_k}^2\ind(E_{M,L}).
\]
Taking unconditional expectations, using the state bound, and iterating from $M$ yields
\[
 \Exp\{\norm{v_k}^2\ind(E_{M,L})\}
 \le C_{M,L}k^{-\lambda}+Ck^{-\alpha}.
\]
Only a deterministic inequality has been multiplied by the indicator, so no martingale centering conditional on $E_{M,L}$ is required. The remainder bound now gives
\[
 \Exp\left\{\ind(E_{M,L})\frac1{\sqrt n}
           \sum_{k=M+1}^n k\norm{\Delta_k}\right\}
 \le C n^{1/2-\alpha}+C_{M,L}n^{1/2-\lambda}=o(1).
\]
On $E_{M,L}$, the prefix through $M$ has a finite numerator and is negligible after division by $\sqrt n$. Let $n\to\infty$, then $L\to\infty$ and $M\to\infty$. Since $\tau$ and all prefix coordinates are finite almost surely, the excluded probability tends to zero. This proves \eqref{eq:finite-prefix-average}. The same proof under the joint law gives the weighted result. For a jointly negligible remainder $R_n^\star$ and $\eps,\delta>0$,
\[
 \Prob\{\Prob^\star(\norm{R_n^\star}>\eps)>\delta\}
 \le\delta^{-1}\Prob(\norm{R_n^\star}>\eps)\to0,
\]
which proves the conditional conclusion.
\end{proof}

\begin{lemma}[Change of base point]\label{lem:random-base}
Under \Cref{ass:intrinsic-comparison}, let $a_n,b_n\in T_{x_\star}\M$ satisfy $a_n=O_p(n^{-1/2})$ and
$b_n=O_{\Prob^\star}(n^{-1/2})$ in probability.  Put
$x_n=R_{x_\star}(a_n)$ and $y_n=R_{x_\star}(b_n)$.  Then
\begin{equation}\label{eq:random-base}
  \sqrt n\,\mathcal P_{x_n\to x_\star}R_{x_n}^{-1}(y_n)
  =\sqrt n(b_n-a_n)+o_{\Prob^\star}(1)
  \quad\text{in probability}.
\end{equation}
In particular,
\begin{equation}\label{eq:random-base-target}
  \sqrt n\,\mathcal P_{x_n\to x_\star}R_{x_n}^{-1}(x_\star)
  =-\sqrt n a_n+o_p(1).
\end{equation}
\end{lemma}

\begin{proof}
By \eqref{eq:Gamma-bound},
\[
  \Gamma(a_n,b_n)=b_n-a_n+R_n,
  \qquad
  \norm{R_n}\le C(\norm{a_n}+\norm{b_n})\norm{b_n-a_n}.
\]
The assumed orders imply
$\sqrt n\norm{R_n}=O_{\Prob^\star}(n^{-1/2})$, proving \eqref{eq:random-base}.  Taking $b_n=0$ gives \eqref{eq:random-base-target}.
\end{proof}

\begin{lemma}[One-step coordinate remainder]\label{lem:one-step}
Under \Cref{ass:geometry}, let $q\in\R$, $u\in B_{T_{x_\star}\M}(0,r)$, and let $\psi\in T_{x_\star}\M$ be such that $z=-\eta q\psi$ lies in the coordinate domain.  Then
\begin{equation*}
  \Phi(u,-\eta q\psi)
  =u-\eta q\psi+\epsilon(u,q,\psi,\eta),
\end{equation*}
where
\begin{equation}\label{eq:one-step-rem}
  \norm{\epsilon(u,q,\psi,\eta)}
  \le C\{\eta|q|\norm u^2\norm\psi+\eta^2q^2\norm\psi^2\}.
\end{equation}
\end{lemma}

\begin{proof}
Substitute $z=-\eta q\psi$ into \eqref{eq:Phi-bound}.  The resulting upper bound is exactly \eqref{eq:one-step-rem}.
\end{proof}

\begin{lemma}[Two non-exponential implementation triples]\label{lem:implementation-geometry}
Use Riemannian parallel transport in the comparison maps.
\begin{enumerate}[label=(\roman*),leftmargin=2.2em]
\item On $\mathbb S^q$, let
\[
 R_x^{\mathrm{proj}}(v)=\frac{x+v}{\norm{x+v}},
 \qquad
 (R_x^{\mathrm{proj}})^{-1}(y)=\frac{y}{\ip{x}{y}}-x,
 \quad \ip{x}{y}>0.
\]
\item On $\Gr(r,p)$, for a horizontal matrix $\mathrm{\Delta}$ with $\mathrm{U}^{\trans}\mathrm{\Delta}=0$, let
\[
 R_{[\mathrm{U}]}^{\mathrm{QR}}(\mathrm{\Delta})=[\operatorname{qf}(\mathrm{U}+\mathrm{\Delta})],
 \qquad
 (R_{[\mathrm{U}]}^{\mathrm{QR}})^{-1}([\mathrm{V}])
 =(\mathrm{I}-\mathrm{U}\mathrm{U}^{\trans})\mathrm{V}(\mathrm{U}^{\trans}\mathrm{V})^{-1},
\]
whenever $\mathrm{U}^{\trans}\mathrm{V}$ is invertible. Here $\operatorname{qf}(\mathrm{Z})$ is the orthonormal $\mathrm{Q}$ factor in a thin QR factorization of a full-column-rank matrix $\mathrm{Z}$; the Grassmann point is independent of the column-sign convention.
\end{enumerate}
On sufficiently small compact coordinate balls, each displayed retraction, inverse, and parallel-transport triple satisfies \eqref{eq:A-bound}, \eqref{eq:Gamma-bound}, and \eqref{eq:Phi-bound}.
\end{lemma}

\begin{proof}
For the sphere, $\norm{x+v}=(1+\norm v^2)^{1/2}$ because $v\perp x$, so
$R_x^{\mathrm{proj}}(v)=x+v-\tfrac12\norm v^2x+O(\norm v^3)$.  This agrees with $\Expmap_x(v)$ through second order, and direct substitution verifies the displayed local inverse.  For the Grassmannian, $[\operatorname{qf}(\mathrm{U}+\mathrm{\Delta})]$ equals the polar-retraction subspace
$[(\mathrm{U}+\mathrm{\Delta})(\mathrm{I}+\mathrm{\Delta}^{\trans}\mathrm{\Delta})^{-1/2}]$ exactly.  The latter agrees with the Grassmann exponential through second order \citep{AbsilEtAl2008,AbsilMalick2012}.  If $[\mathrm{V}]=R_{[\mathrm{U}]}^{\mathrm{QR}}(\mathrm{\Delta})$, then $\mathrm{V}=(\mathrm{U}+\mathrm{\Delta})\mathrm{C}$ for an invertible $\mathrm{C}$, and the displayed graph inverse returns $\mathrm{\Delta}$.

It remains to translate second-order agreement into the three stated comparisons.  In normal coordinates at the target, parallel transport has identity differential and no term linear in the base coordinate.  For either smooth second-order triple, the maps in \eqref{eq:A-def}, \eqref{eq:Gamma-def}, and \eqref{eq:Phi-def} therefore have, respectively, first-order parts $(1-t)a+tb$, $b-a$, and $u+z$.  More explicitly, the Jacobi-field expansion of the normal-coordinate logarithm, composed with radial parallel transport, gives
$D_z\Phi(u,0)=I+O(\norm u^2)$
uniformly on a smaller coordinate ball; second-order agreement of the displayed retractions and inverses preserves this derivative estimate.  Taylor's formula in the step coordinate then yields
\[
 \Phi(u,z)-u-z
 =\{D_z\Phi(u,0)-I\}z
  +\int_0^1(1-s)D^2_{zz}\Phi(u,sz)[z,z]\,ds,
\]
and the second derivative is uniformly bounded.  The other Taylor remainders vanish when $t=0$ and when $a=b$.  Uniform bounds on the same compact ball consequently give
\[
 Ct(\norm a^2+\norm b^2),\qquad
 C(\norm a+\norm b)\norm{b-a},\qquad
 C\{\norm u^2\norm z+\norm z^2\},
\]
which are exactly \eqref{eq:A-bound}, \eqref{eq:Gamma-bound}, and \eqref{eq:Phi-bound}.
\end{proof}

\subsubsection{Proof of the Abstract Bootstrap Theorem}\label{app:abstract}

\begin{lemma}[Negligible martingale remainders]\label{lem:mart-rem}
Let $(z_k,\mathcal H_k)$ be a finite-dimensional martingale-difference sequence.  If
\begin{equation*}
  \frac1n\sum_{k=1}^n
  \Exp(\norm{z_k}^2\mid\mathcal H_{k-1})\xrightarrow{p}0,
\end{equation*}
then $n^{-1/2}\sum_{k=1}^nz_k\to_p0$.  The same statement holds under the joint original--multiplier law.
\end{lemma}

\begin{proof}
Choose an orthonormal basis $e_1,\ldots,e_d$ of the state space. For each $n$, define the martingale
$M^{(n)}_{m,j}=n^{-1/2}\sum_{k=1}^m\ip{z_k}{e_j}$ for $0\le m\le n$. Its predictable quadratic variation is bounded by
\[
  \langle M^{(n)}_{\cdot,j}\rangle_n
  \le \frac1n\sum_{k=1}^n
  \Exp(\norm{z_k}^2\mid\mathcal H_{k-1}).
\]
For $a,\gamma>0$, Lenglart's inequality gives
\[
  \Prob(|M^{(n)}_{n,j}|>a)
  \le \frac{\gamma}{a^2}
      +\Prob\{\langle M^{(n)}_{\cdot,j}\rangle_n>\gamma\}.
\]
First let $n\to\infty$, then $\gamma\downarrow0$.  Hence each coordinate converges to zero in probability.  Finite dimensionality gives the vector conclusion.  The same proof applies to the joint filtration.
\end{proof}

\begin{lemma}[Abel expansion for the base path]\label{lem:base-abel}
Under \Cref{ass:abstract-recursion},
\begin{equation}\label{eq:base-abel}
  \frac1{\sqrt n}\sum_{k=1}^nu_{k-1}
  =-H^{-1}\frac1{\sqrt n}\sum_{k=1}^n\xi_k+o_p(1).
\end{equation}
\end{lemma}

\begin{proof}
Rearrange \eqref{eq:abstract-base}:
\begin{equation*}
  Hu_{k-1}
  =\frac{u_{k-1}-u_k}{\eta_k}
   -\xi_k-b_k-m_k-e_k.
\end{equation*}
Sum from $k=1$ to $n$ and divide by $\sqrt n$.  The first term on the right is $o_p(1)$ by \Cref{lem:abel-boundary} and \eqref{eq:state-abstract}.  The bias and drift terms are $o_p(1)$ by \eqref{eq:bias-abstract} and \eqref{eq:drift-rem-abstract}.  The normalized martingale remainder is $o_p(1)$ by \Cref{lem:mart-rem} and \eqref{eq:mart-rem-abstract}.  Multiplication by the fixed inverse $H^{-1}$ proves \eqref{eq:base-abel}.
\end{proof}

\begin{lemma}[Abel expansion for the weighted path]\label{lem:weighted-abel}
Under \Cref{ass:abstract-recursion},
\begin{equation}\label{eq:weighted-abel}
  \frac1{\sqrt n}\sum_{k=1}^nu_{k-1}^\star
  =-H^{-1}\frac1{\sqrt n}\sum_{k=1}^nW_k\xi_k
  +o_{\Prob^\star}(1)
\end{equation}
in probability.
\end{lemma}

\begin{proof}
Rearranging \eqref{eq:abstract-weighted} gives
\begin{equation}\label{eq:weighted-rearranged-abstract}
  Hu_{k-1}^\star
  =\frac{u_{k-1}^\star-u_k^\star}{\eta_k}
   -W_k\xi_k-W_kb_k-m_k^\star-e_k^\star.
\end{equation}
The normalized Abel boundary is negligible under the joint law by \Cref{lem:abel-boundary} and \eqref{eq:state-abstract}.  The martingale and drift remainders are negligible under the joint law by \Cref{lem:mart-rem}, \eqref{eq:mart-rem-abstract}, and \eqref{eq:drift-rem-abstract}.

For the bias, write
\begin{equation*}
  \frac1{\sqrt n}\sum_{k=1}^nW_kb_k
  =\frac1{\sqrt n}\sum_{k=1}^nb_k
   +\frac1{\sqrt n}\sum_{k=1}^n(W_k-1)b_k.
\end{equation*}
The first term is $o_p(1)$.  Conditionally on the complete original stream, the second term has mean zero and second moment
\begin{equation}\label{eq:weighted-bias-var-proof}
  \Exp^\star\norm{\frac1{\sqrt n}\sum_{k=1}^n(W_k-1)b_k}^2
  =\frac1n\sum_{k=1}^n\norm{b_k}^2\xrightarrow{p}0.
\end{equation}
Conditional Chebyshev therefore makes it $o_{\Prob^\star}(1)$ in probability.

All remaining remainders converge to zero under the joint law.  If $R_n^\star\to0$ in joint probability, then for every $\eps>0$,
\[
  \Prob\{\Prob^\star(\norm{R_n^\star}>\eps)>\eps\}
  \le \eps^{-1}\Prob(\norm{R_n^\star}>\eps)\to0.
\]
Thus joint convergence implies conditional convergence in probability.  Multiplying \eqref{eq:weighted-rearranged-abstract} by $H^{-1}$ proves \eqref{eq:weighted-abel}.
\end{proof}

\begin{proof}[Proof of \Cref{thm:joint}]
Apply \Cref{lem:base-abel,lem:weighted-abel}.  By \Cref{lem:average-expansion} when the local formula holds throughout, or by \Cref{lem:finite-prefix-average} under the \hyperref[rem:localization]{finite-exception localization formulation}, together with \eqref{eq:state-abstract},
\begin{align*}
  \sqrt n\left\{R_{x_\star}^{-1}(\bar x_n)
  -\frac1n\sum_{k=1}^nu_k\right\}&=o_p(1),\\
  \sqrt n\left\{R_{x_\star}^{-1}(\bar x_n^\star)
  -\frac1n\sum_{k=1}^nu_k^\star\right\}
  &=o_{\Prob^\star}(1)
  \quad\text{in probability}.
\end{align*}
Changing the summation index from $u_{k-1}$ to $u_k$ contributes only
$(u_n-u_0)/\sqrt n=o_p(1)$, and similarly under the joint law for the weighted path.  Combining these relations with \eqref{eq:base-abel} and \eqref{eq:weighted-abel} proves \eqref{eq:base-linear-new}--\eqref{eq:weighted-linear-new}.
\end{proof}

\subsubsection{Martingale and Conditional Multiplier Central Limit Theorems}\label{app:clt}

\begin{lemma}[Optional covariance and realized Lindeberg condition]\label{lem:optional-cov}
Under \Cref{ass:innovation},
\begin{align}
  \frac1n\sum_{k=1}^n\xi_k\ot\xi_k&\xrightarrow{p}Q,
  \label{eq:empirical-Q}\\
  \frac1n\sum_{k=1}^n\norm{\xi_k}^2
  \ind\{\norm{\xi_k}>\eps\sqrt n\}&\xrightarrow{p}0
  \quad\text{for every }\eps>0.
  \label{eq:realized-lindeberg}
\end{align}
\end{lemma}

\begin{proof}
Set $X_{n,k}=n^{-1/2}\xi_k$.  For fixed $\eps>0$, the nonnegative adapted process
\[
  A_{n,j}=\sum_{k=1}^j\norm{X_{n,k}}^2
  \ind\{\norm{X_{n,k}}>\eps\}
\]
has predictable compensator
\[
  \widetilde A_{n,j}
  =\sum_{k=1}^j
  \Exp[\norm{X_{n,k}}^2\ind\{\norm{X_{n,k}}>\eps\}\mid\F_{k-1}].
\]
By \eqref{eq:mart-lindeberg}, $\widetilde A_{n,n}\to_p0$.  Lenglart's domination inequality for nonnegative submartingales gives, for $a,b>0$, $\Prob(A_{n,n}>a) \le b/a+\Prob(\widetilde A_{n,n}>b)$. Let $n\to\infty$ and then $b\downarrow0$ to obtain \eqref{eq:realized-lindeberg}.

It remains to compare optional and predictable covariances.  Define
\begin{equation*}
  \mathrm{Y}_{n,k}^{(\eps)}
  :=X_{n,k}\ot X_{n,k}\ind\{\norm{X_{n,k}}\le\eps\}
  -\Exp[X_{n,k}\ot X_{n,k}\ind\{\norm{X_{n,k}}\le\eps\}\mid\F_{k-1}].
\end{equation*}
These are matrix-valued martingale differences.  On the event
\[
  \sum_{k=1}^n\Exp(\norm{X_{n,k}}^2\mid\F_{k-1})\le M,
\]
the predictable sum of their squared Frobenius norms is at most $4\eps^2M$.  Coordinatewise Lenglart therefore implies $\sum_{k=1}^n\mathrm{Y}_{n,k}^{(\eps)}=O_p(\eps)$ after first restricting to the preceding event.  The event has probability tending to one for sufficiently large $M$ by \eqref{eq:predictable-Q}.  The omitted large-increment parts are $o_p(1)$ by \eqref{eq:mart-lindeberg}, \eqref{eq:realized-lindeberg}, and another application of Lenglart.  Letting $\eps\downarrow0$ yields
\[
  \sum_{k=1}^nX_{n,k}\ot X_{n,k}
  -\sum_{k=1}^n\Exp(X_{n,k}\ot X_{n,k}\mid\F_{k-1})
  \xrightarrow{p}0.
\]
Combining this with \eqref{eq:predictable-Q} proves \eqref{eq:empirical-Q}.
\end{proof}

\begin{lemma}[Base martingale central limit theorem]\label{lem:base-clt}
Under \Cref{ass:innovation},
\begin{equation}\label{eq:base-mart-clt}
  \frac1{\sqrt n}\sum_{k=1}^n\xi_k
  \rightsquigarrow N(0,Q).
\end{equation}
\end{lemma}

\begin{proof}
For every fixed $a\in T_{x_\star}\M$, the scalar martingale array
$n^{-1/2}\ip{a}{\xi_k}$ has predictable variance converging to
$\ip{a}{Qa}$ by \eqref{eq:predictable-Q}.  Its conditional Lindeberg condition follows from \eqref{eq:mart-lindeberg} because
$|\ip{a}{\xi_k}|\le\norm a\norm{\xi_k}$.  The scalar martingale central limit theorem \citep[Chapter~3]{HallHeyde1980} gives convergence of every linear functional.  Cram\'er--Wold and finite dimensionality prove \eqref{eq:base-mart-clt}.
\end{proof}

\begin{lemma}[Conditional multiplier central limit theorem with second moments]\label{lem:multiplier-clt}
Under \Cref{ass:innovation},
\begin{equation}\label{eq:multiplier-clt}
  d_{\BL}\left[
  \cL^\star\left\{\frac1{\sqrt n}\sum_{k=1}^n(W_k-1)\xi_k\right\},
  N(0,Q)
  \right]\xrightarrow{p}0.
\end{equation}
\end{lemma}

\begin{proof}
Condition on the realized sequence $\xi_1,\ldots,\xi_n$.  The summands are independent, centered, and their conditional covariance is
$n^{-1}\sum_{k=1}^n\xi_k\ot\xi_k$, which converges to $Q$ in probability by \Cref{lem:optional-cov}.

It remains to verify the conditional Lindeberg condition.  Put $U=|W-1|$.  For $M>0$ and $\eps>0$,
\begin{align*}
 &\frac1n\sum_{k=1}^n
 \Exp^\star\left[U_k^2\norm{\xi_k}^2
 \ind\{U_k\norm{\xi_k}>\eps\sqrt n\}\right]
 \\
 &\quad\le
 \Exp\{U^2\ind(U>M)\}\frac1n\sum_{k=1}^n\norm{\xi_k}^2
 \\
 &\qquad
 +\Exp(U^2)\frac1n\sum_{k=1}^n\norm{\xi_k}^2
 \ind\{\norm{\xi_k}>\eps\sqrt n/M\}.
\end{align*}
The empirical second moment is $O_p(1)$ by \eqref{eq:empirical-Q}.  Since $\Exp U^2<\infty$, the first coefficient tends to zero as $M\to\infty$.  For fixed $M$, the second line tends to zero in probability by \eqref{eq:realized-lindeberg}.  Hence the conditional Lindeberg condition holds in probability.

The conditional Lindeberg--Feller theorem now gives convergence of every conditional linear functional along every subsequence on which the covariance and Lindeberg conditions hold almost surely.  Every subsequence has a further subsequence with this property, so the convergence holds in probability.  Finite dimensionality upgrades Cram\'er--Wold convergence to bounded-Lipschitz convergence, proving \eqref{eq:multiplier-clt}.
\end{proof}

\begin{proof}[Proof of \Cref{thm:bootstrap}]
The change-of-base lemma requires conditional root-$n$ tightness of the weighted average.  From \eqref{eq:weighted-linear-new},
\[
 \sqrt nR_{x_\star}^{-1}(\bar x_n^\star)
 =-H^{-1}\left\{\frac1{\sqrt n}\sum_{k=1}^n\xi_k
 +\frac1{\sqrt n}\sum_{k=1}^n(W_k-1)\xi_k\right\}
 +o_{\Prob^\star}(1).
\]
The first sum is $O_p(1)$ by \Cref{lem:base-clt}.  Conditionally on the original stream, the second sum has second moment $n^{-1}\sum_k\norm{\xi_k}^2=O_p(1)$ by \Cref{lem:optional-cov}; conditional Markov inequality therefore makes it $O_{\Prob^\star}(1)$ in probability.  Hence $R_{x_\star}^{-1}(\bar x_n^\star)=O_{\Prob^\star}(n^{-1/2})$, as required.

Apply \Cref{thm:joint} and \Cref{lem:random-base} with
$a_n=R_{x_\star}^{-1}(\bar x_n)$ and
$b_n=R_{x_\star}^{-1}(\bar x_n^\star)$.  On the event that the local inverse is available, subtracting \eqref{eq:base-linear-new} from \eqref{eq:weighted-linear-new} and using \Cref{lem:random-base} gives \eqref{eq:bootstrap-linear-new}.  The conditional probability of the complementary event tends to zero in probability by the root-$n$ tightness just established and the local inverse domain, so the measurable completion in \eqref{eq:Tstar-completed} preserves the expansion.  \Cref{lem:multiplier-clt} and conditional Slutsky yield \eqref{eq:bootstrap-BL-new}.  By \Cref{lem:base-clt} and \eqref{eq:base-linear-new}, the sampling law of $\sqrt nR_{x_\star}^{-1}(\bar x_n)$ converges to $N(0,V)$.  The triangle inequality for $d_{\BL}$ proves \eqref{eq:bootstrap-target-new}.
\end{proof}

\begin{proof}[Proof of \Cref{thm:fixedB}]
Let $f_0,\ldots,f_B$ be bounded continuous functions on $T_{x_\star}\M$.  Conditional on the original stream, the $B$ multiplier paths are independent and identically distributed.  Hence
\begin{align*}
 &\Exp\left[f_0(Z_n)\prod_{b=1}^B f_b(T_{n,b}^\star)\right]\\
 &\qquad=\Exp\left[f_0(Z_n)
       \prod_{b=1}^B\Exp^\star f_b(T_n^\star)\right].
\end{align*}
By \eqref{eq:bootstrap-BL-new}, for each $b$,
$\Exp^\star f_b(T_n^\star)\to_p\Exp f_b(Z_b)$, where
$Z_b\sim N(0,V)$.  Boundedness makes this convergence hold in $L^1$, and therefore the product of the conditional expectations converges in $L^1$ to
$\prod_{b=1}^B\Exp f_b(Z_b)$.  By \Cref{lem:base-clt} and
\eqref{eq:base-linear-new}, $Z_n\rightsquigarrow Z_0\sim N(0,V)$.  Consequently,
\[
 \Exp\left[f_0(Z_n)\prod_{b=1}^B f_b(T_{n,b}^\star)\right]
 \longrightarrow
 \prod_{b=0}^B\Exp f_b(Z_b).
\]
Finite products of bounded continuous functions form a convergence-determining class on the finite product space.  This proves \eqref{eq:fixedB-joint}.
\end{proof}

\begin{proof}[Proof of \Cref{cor:fixedB}]
For the scalar statement, differentiability and \Cref{lem:random-base} give jointly for $b=1,\ldots,B$,
\begin{align*}
 \sqrt n\{\phi(\bar x_n)-\phi(x_\star)\}
   &=D\phi_{x_\star}Z_n+o_p(1),\\
 \sqrt n\{\phi(\bar x_n^{\star(b)})-\phi(\bar x_n)\}
   &=D\phi_{x_\star}T_{n,b}^\star+o_p(1).
\end{align*}
By \Cref{thm:fixedB}, after division by $\sigma_\phi>0$, the first term and the $B$ bootstrap differences converge to independent standard normal variables $N_0,N_1,\ldots,N_B$.  Therefore
\[
 \frac{\phi(\bar x_n)-\phi(x_\star)}{s_{\phi,n,B}}
 \rightsquigarrow
 \frac{N_0}{\{B^{-1}\sum_{b=1}^BN_b^2\}^{1/2}}
 \sim t_B.
\]
The limiting denominator is positive almost surely, so $\Prob(s_{\phi,n,B}=0)\to0$.  The Student quantile is continuous, and the fallback convention is asymptotically irrelevant, which proves \eqref{eq:fixedB-CI-coverage}.

For the intrinsic region, transport $\widehat S_{n,B}$ from $T_{\bar x_n}\M$ to $T_{x_\star}\M$ and identify the latter with $\R^d$ through an orthonormal basis. In the following Gaussian calculation, $\mathrm{S}_B$ and $\mathrm{V}$ denote the corresponding covariance matrices.  By \Cref{thm:fixedB} and the continuous mapping theorem,
\[
 \mathcal P_{\bar x_n\to x_\star}\widehat S_{n,B}
 \mathcal P_{x_\star\to\bar x_n}
 \rightsquigarrow
 \mathrm{S}_B:=\frac1B\sum_{b=1}^BZ_b\ot Z_b.
\]
Since $\mathrm{V}$ is positive definite and $B\ge d$, $\mathrm{S}_B$ is invertible almost surely.  Also, \Cref{lem:random-base} gives
\[
 \sqrt n\,\mathcal P_{\bar x_n\to x_\star}R_{\bar x_n}^{-1}(x_\star)
 =-Z_n+o_p(1).
\]
Thus the quadratic form determining membership of $x_\star$ in \eqref{eq:fixedB-region} converges to $Z_0^\trans \mathrm{S}_B^{-1}Z_0$. Since $B\mathrm{S}_B=\sum_{b=1}^BZ_bZ_b^\trans$,
$B\mathrm{S}_B\sim\mathsf W_d(\mathrm{V},B)$, the Wishart law with scale $\mathrm{V}$ and $B$ degrees of freedom, and is independent of $Z_0\sim N_d(0,\mathrm{V})$.  The standard normal--Wishart identity gives
\[
 \frac{B-d+1}{dB}Z_0^\trans \mathrm{S}_B^{-1}Z_0
 \sim F_{d,B-d+1}.
\]
The limiting inverse-retraction vector lies in the regular local domain and $\mathrm{S}_B$ is nonsingular almost surely; hence the finite-sample fallback event has probability tending to zero.  The distribution is continuous, so \eqref{eq:fixedB-critical} gives \eqref{eq:fixedB-region-coverage}.
\end{proof}

\begin{proof}[Proof of \Cref{prop:euclidean-reduction}]
In Euclidean space, $R_x(v)=x+v$, inverse retraction is subtraction, parallel transport is the identity, and \eqref{eq:base-average} is the arithmetic mean.  Equations \eqref{eq:base-update} and \eqref{eq:weighted-update} therefore become
\[
 x_k=x_{k-1}-\eta_k\nabla h(x_{k-1},Z_k),
 \qquad
 x_k^{\star(b)}=x_{k-1}^{\star(b)}-
 \eta_kW_k^{(b)}\nabla h(x_{k-1}^{\star(b)},Z_k),
\]
which are the original and perturbed ASGD paths in COnB.  For $\phi_i(x)=x_i$, Equation \eqref{eq:fixedB-scale} is the uncentered squared coordinate displacement from the observed averaged path, and \eqref{eq:fixedB-CI} uses the $t_B$ critical value.  These are exactly the COnB ASGD coordinate interval after matching the update index.  The final statement follows from \Cref{thm:fixedB,cor:fixedB}.
\end{proof}

\begin{proof}[Proof of \Cref{cor:functionals}]
Differentiability at $x_\star$ gives, uniformly for $h\to0$ in the local chart,
\[
  \phi\{R_{x_\star}(h)\}-\phi(x_\star)
  =D\phi_{x_\star}h+o(\norm h).
\]
The two averages are $O_p(n^{-1/2})$ and $O_{\Prob^\star}(n^{-1/2})$ in fixed coordinates by \Cref{thm:joint}.  Subtracting their two expansions therefore yields
\[
  \sqrt n\{\phi(\bar x_n^\star)-\phi(\bar x_n)\}
  =D\phi_{x_\star}T_n^\star+o_{\Prob^\star}(1),
\]
where \Cref{lem:random-base} aligns the tangent difference.  Conditional Slutsky proves \eqref{eq:functional-limit}.

For scalar basic intervals, conditional weak convergence to a continuous nondegenerate Gaussian distribution implies convergence of the relevant conditional quantiles.  The empirical bootstrap quantiles with $B_n\to\infty$ converge to the conditional quantiles by the conditional Glivenko--Cantelli theorem.  The intrinsic-region statement follows in the same way from continuity of the norm quantile and the isometry of parallel transport.
\end{proof}

\begin{proof}[Proof of \Cref{prop:covariance}]
On the original-stream event $\{\bar x_{n}\in U_{\star}\}$, whose probability tends to one, $T_n^\star=\mathcal P_{\bar x_n\to x_\star}\widetilde T_n^\star$ and parallel transport is an isometry. Thus conditional weak convergence and conditional uniform integrability of $\norm{T_n^\star}^2$ imply, by conditional truncation,
\begin{align}
  \Exp^\star T_n^\star&\xrightarrow{p}0,
  \label{eq:cond-first-moment}\\
  \Exp^\star(T_n^\star\ot T_n^\star)&\xrightarrow{p}V.
  \label{eq:cond-second-moment}
\end{align}
For completeness, take a bounded continuous radial truncation $h_K(t)$ equal to $t$ on $\norm t\le K$.  Conditional weak convergence gives convergence of the first two moments of $h_K(T_n^\star)$.  The conditional uniform-integrability condition makes the truncation error vanish in probability as $K\to\infty$, which proves \eqref{eq:cond-first-moment}--\eqref{eq:cond-second-moment}.

Given the original stream, the $B_n$ bootstrap replicates are i.i.d. from the conditional law.  For fixed $K$, the sample first and second moments of the truncated variables differ from their conditional expectations by $o_p(1)$ because their conditional variances are bounded by constants times $K^2/B_n$ and $K^4/B_n$.  The average tail contribution is controlled by conditional Markov inequality and uniform integrability.  Letting first $n\to\infty$ and then $K\to\infty$ shows that the transported empirical first and second moments converge to \eqref{eq:cond-first-moment}--\eqref{eq:cond-second-moment}.  Their centered difference is therefore $V+o_p(1)$, proving \eqref{eq:cov-consistency}.

If $V$ is positive definite, inversion is continuous in a neighborhood of $V$.  By \eqref{eq:random-base-target}, the transported tangent vector from $\bar x_n$ to $x_\star$ has asymptotic law $N(0,V)$ after multiplication by $\sqrt n$.  Slutsky's theorem yields a $\chi_d^2$ limit for the Wald quadratic form and hence the coverage claim.
\end{proof}

\subsubsection{Verification of the Primitive Conditions and State Bounds}\label{app:primitive}

\begin{proof}[Proof of \Cref{prop:primitive-verification}]
Choose a radius $a>0$ within the oracle and comparison domains whose ball contains the eventual localization ball. For every $k$, define the predictable representatives $\widehat u_{k-1}=u_{k-1}\ind\{\norm{u_{k-1}}<a\}$ and $\widehat u_{k-1}^\star=u_{k-1}^\star\ind\{\norm{u_{k-1}^\star}<a\}$. Define, at every index,
\[
 m_k=\chi_k(\widehat u_{k-1}),\qquad
 m_k^\star=W_k\chi_k(\widehat u_{k-1}^\star)
                 +(W_k-1)H\widehat u_{k-1}^\star.
\]
These are martingale differences by \eqref{eq:chi-centered} and independence of the current weight from the original stream and past weights. The oracle need not be independent of the past. Define $e_{k},e_{k}^{\star}$ by the exact residual identities
\begin{align*}
 e_{k}&=\eta_{k}^{-1}(u_{k-1}-u_k)-Hu_{k-1}-\xi_k-b_k-m_k,\\
 e_{k}^{\star}&=\eta_{k}^{-1}(u_{k-1}^\star-u_k^\star)
       -Hu_{k-1}^{\star}-W_k\xi_k-W_kb_k-m_k^\star.
\end{align*}
All arrays are now defined globally, and \eqref{eq:abstract-base}--\eqref{eq:abstract-weighted} hold identically. No martingale array is truncated at a last-exception time or by a current-proposal event.

The conditional moment bound at predictable local arguments gives, for a sufficiently small fixed $r_1>0$,
\[
 \Prob\{\eta_k\norm{\psi_k(\widehat u_{k-1})}>r_1\mid\F_{k-1}\}
 +\Prob\{\eta_k|W_k|\norm{\psi_k(\widehat u_{k-1}^\star)}>r_1\mid\G_{k-1}\}
 \le C\eta_k^2.
\]
Since $\sum_{k}\eta_{k}^2<\infty$, only finitely many such large proposals occur almost surely. By the assumed eventual chart entry and safeguard inactivity, each path has only finitely many indices at which its representative differs from its actual state or the local update formula fails. These are its exceptional indices, separately for each multiplier path.

At regular indices, \Cref{lem:one-step} gives
\begin{equation}\label{eq:coordinate-base-primitive}
 u_k=u_{k-1}-\eta_k\psi_k(u_{k-1})+\epsilon_k.
\end{equation}
Define $\epsilon_{k}$ to be this coordinate error at regular indices and zero at exceptional ones. Then, at all indices,
\[
 \norm{\epsilon_k}\le C\{\eta_k\norm{\widehat u_{k-1}}^2
       \norm{\psi_k(\widehat u_{k-1})}
       +\eta_k^2\norm{\psi_k(\widehat u_{k-1})}^2\}.
\]
Define $\epsilon_k^\star$ in the same way for the weighted local update. Its bound is
\begin{equation}\label{eq:eps-weighted-primitive}
 \norm{\epsilon_k^\star}\le C\{\eta_k|W_k|\norm{\widehat u_{k-1}^\star}^2
       \norm{\psi_k(\widehat u_{k-1}^\star)}
       +\eta_k^2W_k^2\norm{\psi_k(\widehat u_{k-1}^\star)}^2\}.
\end{equation}
Substituting \eqref{eq:primitive-expansion} into \eqref{eq:coordinate-base-primitive} shows that $e_{k}$ agrees with $r_{\mathrm{pop}}(\widehat u_{k-1})-\eta_k^{-1}\epsilon_k$ at every regular index. Likewise, $e_k^\star$ agrees there with $W_kr_{\mathrm{pop}}(\widehat u_{k-1}^\star)-\eta_k^{-1}\epsilon_k^\star$. Their differences at the exceptional indices have an almost surely finite sum, because all actual coordinates and residuals are finite. Dividing that sum by $\sqrt n$ gives $o_{p}(1)$ under the original or joint law. This use of finite exceptions does not require them to form stopping times.

For the martingale variance, \eqref{eq:chi-L2} gives
\[
 \Exp(\norm{m_k}^2\mid\F_{k-1})\le\omega(\norm{\widehat u_{k-1}}),\qquad
 \Exp(\norm{m_k^\star}^2\mid\G_{k-1})
 \le C\{\omega(\norm{\widehat u_{k-1}^\star})+\norm{\widehat u_{k-1}^\star}^2\}.
\]
The representatives converge to zero in probability by \eqref{eq:state-abstract} and satisfy $\norm{\widehat u}\le\norm u$. Boundedness of $\omega$ and $\omega(t)\to0$ imply that the expectations of these bounds tend to zero. Ces\`aro averaging and Markov's inequality establish \eqref{eq:mart-rem-abstract}, under the joint law for the weighted path.

For the population drift, Jensen's inequality and the state bound imply
\begin{equation}\label{eq:r0-sum-proof}
 \frac1{\sqrt n}\sum_{k=1}^n\Exp\norm{r_{\mathrm{pop}}(\widehat u_{k-1})}
 \le\frac C{\sqrt n}\left\{1+\sum_{k=2}^{n}\eta_{k-1}^{(1+\rho)/2}\right\}=o(1)
\end{equation}
by \eqref{eq:alpha-rho}. The weighted counterpart follows using $\Exp|W|<\infty$ and the joint state bound. Conditional Cauchy--Schwarz at the predictable representatives gives $\Exp\norm{\eta_k^{-1}\epsilon_k}\le C\eta_{k}$ for $k\ge2$, and \eqref{eq:eps-weighted-primitive} gives $\Exp\Exp^\star\norm{\eta_k^{-1}\epsilon_k^\star}\le C\eta_k$. Here only $\Exp W^2=2$ and the conditional oracle second moment are used. These bounds hold without conditioning on a non-stopping regularity event, since the errors are bounded by nonnegative envelopes at all indices. Their root-$n$ sums vanish because $\alpha>1/2$; together with \eqref{eq:r0-sum-proof} and the finite exceptional residuals, this proves \eqref{eq:drift-rem-abstract}.

The bias conditions are assumed, and \eqref{eq:weighted-bias-var-proof} controls their multiplier fluctuation. Thus \Cref{ass:abstract-recursion} holds. All constructions are separate for each weighted path and retain conditional i.i.d. sampling. The complete averages keep their exceptional prefixes, whose first-order negligibility follows from \Cref{lem:finite-prefix-average}.
\end{proof}

\begin{proof}[Proof of \Cref{prop:state-rate}]
Write $\mathcal V(x)=d(x,x_\star)^2$.  Because $K$ is compact and $\mathcal V$ is $C^2$ on $K^+$, there are constants $r_0,C>0$ such that, for $x\in K$ and every tangent vector $v$ with $\norm v_x\le r_0$,
\begin{equation}\label{eq:distance-taylor}
  \mathcal V\{R_x(-v)\}
  \le \mathcal V(x)+2\ip{v}{\Log_x(x_\star)}_x+C\norm v_x^2.
\end{equation}
Decrease the safeguard radius $r_0$, if necessary, so that every short retraction proposal remains in $K^+$.  Let $q=1$ for the base path and $q=W_k$ for a weighted path, and abbreviate
$\Psi=\Psi_k(x;D_k)$, $\eta=\eta_k$.  Define the short-proposal event $A=\{\eta|q|\norm\Psi_x\le r_0\}$. On $A$, apply \eqref{eq:distance-taylor} with $v=\eta q\Psi$.  On $A^c$, the safeguarded next state lies in $K$, so its squared distance is bounded by the squared diameter $D_K^2$.

We first control the discarded tail in the drift.  With $X=|q|\norm\Psi_x$,
\begin{equation}\label{eq:tail-first-moment}
  \Exp\{X\ind(\eta X>r_0)\mid\text{past}\}
  \le \frac{\eta}{r_0}\Exp(X^2\mid\text{past}),
\end{equation}
and
\begin{equation}\label{eq:tail-prob}
  \Prob(\eta X>r_0\mid\text{past})
  \le \frac{\eta^2}{r_0^2}\Exp(X^2\mid\text{past}).
\end{equation}
For the base path, \eqref{eq:oracle-second-state} bounds the conditional second moment of $X$.  For a weighted path, independence and \eqref{eq:weight-second} give the same bound multiplied by $\Exp W^2=2$.

Taking conditional expectations in the short-event Taylor inequality, restoring the omitted drift by \eqref{eq:tail-first-moment}, and controlling the long event by \eqref{eq:tail-prob} gives, for either path,
\begin{equation}\label{eq:state-pre-drift}
  \Exp\{\mathcal V(x_k)\mid\text{past}\}
  \le \mathcal V(x)+2\eta\ip{\bar G_k(x)}{\Log_x(x_\star)}_x+C\eta^2,
\end{equation}
where $\bar G_k(x)=\Exp\{\Psi_k(x;D_k)\mid\F_{k-1}\}$; for the weighted path we used
$\Exp(W_k\Psi_k\mid\text{past})=\bar G_k$.
By \eqref{eq:strong-monotonicity}--\eqref{eq:mean-bias-state},
\begin{align*}
  \ip{\bar G_k(x)}{\Log_x(x_\star)}_x
  &\le-\mu\mathcal V(x)+a_k\sqrt{\mathcal V(x)}\\
  &\le-\frac\mu2\mathcal V(x)+\frac{a_k^2}{2\mu}.
\end{align*}
Since $a_k^2\le C\eta_k$, \eqref{eq:state-pre-drift} becomes
\begin{equation}\label{eq:state-recursion}
  \Exp\{\mathcal V(x_k)\mid\text{past}\}
  \le(1-c\eta_k)\mathcal V(x_{k-1})+C\eta_k^2
\end{equation}
for some $c>0$.  The safeguard does not increase squared distance on short proposals and is already accounted for by the bounded long-event contribution.

Taking expectations, let $A_k=\Exp\mathcal V(x_k)$ for the base path or $A_k=\Exp\Exp^\star\mathcal V(x_k^\star)$ for the weighted path.  Then $A_k\le(1-c\eta_k)A_{k-1}+C\eta_k^2$. Because $\eta_{k-1}-\eta_k=o(\eta_k^2)$ when $\alpha<1$, a comparison induction proves $A_k\le C_0\eta_k$ for a sufficiently large $C_0$.  Indeed, if $A_{k-1}\le C_0\eta_{k-1}$, then
\[
  A_k-C_0\eta_k
  \le C_0(\eta_{k-1}-\eta_k)-cC_0\eta_k\eta_{k-1}+C\eta_k^2,
\]
and the negative middle term dominates the first and third terms for large $k$ after $C_0$ is chosen large enough.  This proves \eqref{eq:state-rate-conclusion}.

For almost sure inactivity, \eqref{eq:state-recursion}, $\sum\eta_k=\infty$, and $\sum\eta_k^2<\infty$ allow the Robbins--Siegmund theorem to be applied.  It gives convergence of $\mathcal V(x_k)$ and $\sum\eta_k\mathcal V(x_{k-1})<\infty$ almost surely, hence $\mathcal V(x_k)\to0$.  State convergence puts the path in $U_0$ eventually.  Apply the same tail estimate as \eqref{eq:tail-prob} at the smaller uniform identity threshold $\delta_0$ from \eqref{eq:safeguard-identity}:
\[
  \sum_{k=1}^\infty
  \Prob(\eta_k|q_k|\norm{\Psi_k}>\delta_0\mid\text{past})<\infty
  \quad\text{almost surely}.
\]
Conditional Borel--Cantelli implies that only finitely many proposals exceed $\delta_0$.  Therefore, after a finite random index, the state lies in $U_0$ and every proposal meets \eqref{eq:safeguard-identity}; the safeguarded update then equals the original retraction update forever.
\end{proof}

\begin{proof}[Proof of \Cref{lem:averaging-basin}]
Invariance and induction give $\bar x_k\in K$.  Put
$a_k=d(\bar x_k,x_\star)^2$ and $b_k=d(x_k,x_\star)^2$.  Then \eqref{eq:averaging-basin-contraction} gives
\begin{equation}\label{eq:averaging-basin-scalar}
  a_k\le\left(1-\frac\kappa k\right)a_{k-1}+\frac Ck b_k.
\end{equation}
For fixed $m<k$, let
\[
  p_{j,k}=\prod_{\ell=j+1}^k\left(1-\frac\kappa\ell\right).
\]
Iteration of \eqref{eq:averaging-basin-scalar} yields
\[
  a_k\le p_{m,k}a_m+C\sum_{j=m+1}^k\frac{p_{j,k}}j b_j.
\]
Here $p_{m,k}\to0$ and
\[
  \sum_{j=m+1}^k\frac{p_{j,k}}j
  =\frac{1-p_{m,k}}\kappa.
\]
Given $\eps>0$, choose $m$ so that $b_j\le\eps$ for $j\ge m$.  It follows that
$\limsup_k a_k\le C\eps/\kappa$; letting $\eps\downarrow0$ proves $\bar x_k\to x_\star$.

For the moment bound, take expectations in \eqref{eq:averaging-basin-scalar}.  Since $p_{j,k}\le C_2(j/k)^\kappa$,
\[
  \Exp a_k
  \le p_{1,k}\Exp a_1
     +C_3\sum_{j=2}^k p_{j,k}j^{-1-\alpha}
  \le C_4k^{-\kappa}
     \left(1+\sum_{j=2}^k j^{\kappa-\alpha-1}\right)
  =O(k^{-\alpha}),
\]
where the last equality uses $\kappa>\alpha$.  The argument is deterministic before expectations are taken, so replacing $\Exp$ by $\Exp\Exp^\star$ proves the joint-law statement.
\end{proof}

\subsubsection{Proofs for the Non-Euclidean Applications}\label{app:applications}

\begin{proof}[Proof of \Cref{cor:sphere}]
For $f_y(x)=\tfrac12d(x,y)^2$ away from the cut locus,
$\grad f_y(x)=-\Log_x(y)$.  Hence the target innovation in fixed coordinates is
$\xi=-v$, where $v=\Log_{x_\star}(Y)$, and it is centered by
\eqref{eq:sphere-support}.  Its covariance is $Q_{\mathbb S}$ in \eqref{eq:sphere-Q}.

The Hessian of $f_y$ at $x_\star$ on the unit sphere is
\[
 \Hess f_y(x_\star)
 =e\ot e+r\cot(r)\{I-e\ot e\},
\]
with the continuous convention at $r=0$.  Taking expectations gives
$H_{\mathbb S}$.  Since $r\le r_0<\pi/2$, both eigenvalues $1$ and
$r\cot r$ are strictly positive, so $H_{\mathbb S}$ is positive definite.

On a common strongly convex neighborhood containing the support and localized paths, the log map and its derivatives are bounded.  The stochastic score is therefore square integrable, locally Lipschitz in $L^2$, and its population field has a quadratic remainder.  The exponential map and parallel transport satisfy the geometric assumptions.  The state-rate hypothesis and \Cref{prop:primitive-verification} yield the abstract iteration with $\rho=1$.  Applying \Cref{thm:joint} and using $-H_{\mathbb S}^{-1}\xi=H_{\mathbb S}^{-1}v$ gives \eqref{eq:sphere-limit}.  \Cref{thm:bootstrap,cor:fixedB,cor:functionals} give the inferential statements.
\end{proof}

\begin{proof}[Proof of \Cref{cor:sphere-averaging-basin}]
The cap $K_\rho$ is strongly geodesically convex, and squared distance from $x_\star$ is geodesically convex on it.  Exponential averaging follows the minimizing geodesic segment and therefore satisfies \eqref{eq:averaging-basin-contraction} with $\kappa=C=1$.

For normalized projection, take $x,y\in K_\rho$, write $\theta=d(x,y)\le2\rho<\pi/2$, and express $y=\cos(\theta)x+\sin(\theta)u$ with $u\perp x$ and $\norm u=1$.  For
\[
  R_x^{\mathrm{proj}}(v)=\frac{x+v}{\norm{x+v}},
  \qquad
  (R_x^{\mathrm{proj}})^{-1}(y)=\frac{y}{\ip{x}{y}}-x,
\]
direct substitution gives
\[
  R_x^{\mathrm{proj}}\{t(R_x^{\mathrm{proj}})^{-1}(y)\}
  =\Expmap_x\{s(t,\theta)\Log_x(y)\},
  \qquad
  s(t,\theta)=\frac{\arctan\{t\tan(\theta)\}}{\theta},
\]
with the continuous convention at $\theta=0$.  Concavity of $t\mapsto\arctan\{t\tan(\theta)\}$ gives $s(t,\theta)\ge t$, while monotonicity of $\tan(\theta)/\theta$ gives
\[
  s(t,\theta)\le \frac{\tan(2\rho)}{2\rho}t=:C_\rho t.
\]
Moreover, $0\le s(t,\theta)\le1$, so the update lies on the same minimizing geodesic segment and remains in $K_\rho$.  Geodesic convexity now yields
\begin{align*}
 d\{\mathsf A_t(x,y),x_\star\}^2
 &\le(1-s)d(x,x_\star)^2+s d(y,x_\star)^2\\
 &\le(1-t)d(x,x_\star)^2+C_\rho t d(y,x_\star)^2.
\end{align*}
Thus \Cref{lem:averaging-basin} applies with $\kappa=1>\alpha$ to the base path and, under the joint law, separately to every fixed multiplier path.  Convergence then gives the path-specific entry times into the comparison chart.
\end{proof}

\begin{proof}[Proof of \Cref{cor:hadamard}]
A Hadamard manifold has no cut locus, so $f_y(x)=\tfrac12d(x,y)^2$ is smooth and
$\grad f_y(x)=-\Log_x(y)$ globally.  At the minimizer,
$\Exp\Log_{x_\star}(Y)=0$.  Thus the target innovation is
$\xi=-\Log_{x_\star}(Y)$ with covariance $Q_F$, and the derivative of the population field is
$H_F=\Hess F(x_\star)$.

Compact support and compact localization make the score, its local derivatives, and the relevant curvature terms bounded.  They imply the local $L^2$ modulus, a quadratic population remainder, and the oracle second-moment envelope.  Strong geodesic convexity of squared distance supplies local monotonicity, while the state rate is assumed in the corollary.  The primitive theorem therefore applies with $\rho=1$.  Since $H_F$ is self-adjoint, \Cref{thm:joint} gives \eqref{eq:hadamard-limit}, and \Cref{thm:bootstrap,cor:fixedB,cor:functionals} give conditional and fixed-replication inference.

For the positive-definite specialization, formulas \eqref{eq:spd-exp}--\eqref{eq:spd-log} are the affine-invariant exponential and logarithm.  An eigenvalue-bounded set is compact in finite dimension and stays a positive distance from the boundary, so the preceding local bounds apply.  Finally, the usual matrix differential gives
$D\log\det(\mathrm{X})[\mathrm{U}]=\tr(\mathrm{X}^{-1}\mathrm{U})$, proving \eqref{eq:logdet-derivative}.
\end{proof}

\begin{proof}[Proof of \Cref{cor:spd-lognormal}]
Affine congruence by $\mathrm{A}^{-1/2}$ is an isometry, and \eqref{eq:spd-log} gives
\[
  \mathrm{A}^{-1/2}\Log_{\mathrm{A}}(\mathrm{Y})\mathrm{A}^{-1/2}=\mathrm{E},
  \qquad d(\mathrm{A},\mathrm{Y})=\norm{\mathrm{E}}_F.
\]
Sign symmetry therefore centers the mean score at $\mathrm{A}$.  Finite second distance moment gives existence, and strict geodesic convexity on the Hadamard manifold gives uniqueness.  On any compact set of states, Hessian comparison for the affine-invariant metric bounds the transported score and its local derivative by $C_K(1+\norm{\mathrm{E}}_F)$.  The assumed $2+\delta$ moment consequently supplies the score second moment, local $L^2$ modulus, and oracle envelope directly, without compact support for $\mathrm{Y}$.  The quantitative population remainder and its step-size threshold are separate hypotheses of the corollary; with them, the stated localization, and the state rates, \Cref{prop:primitive-verification} applies.

For the Gaussian contamination model in the numerical experiments, the map $s_{\mathrm{A}}(\mathrm{Y})=\mathrm{A}\mathrm{Y}^{-1}\mathrm{A}$ is an affine-invariant isometry and sends $\mathrm{E}$ to $-\mathrm{E}$. The independent symmetric contamination preserves this invariance and finite distance moments. The nondegenerate Gaussian component gives a positive density on $\SPD_m$, so the law is not supported on a geodesic. Convexity of distance on this Hadamard manifold is then strict after expectation along every nonconstant geodesic, giving a unique median. Isometry invariance fixes both unique minimizers under $s_{\mathrm{A}}$, whose only positive-definite fixed point is $\mathrm{A}$. This establishes the common target for the simulated laws; the path and regularity conditions for inference remain those stated in the corollary.

For the median, the normalized score at $\mathrm{A}$ is $-\mathrm{E}/\norm{\mathrm{E}}_F$ under the same isometry, with any bounded convention at $\mathrm{E}=0$.  Its expectation is zero by sign symmetry.  The supremum-envelope argument in the proof of \Cref{cor:median} supplies the local $L^2$ modulus. Together with the median-specific localization and state-rate assumptions and its stated population regularity, this verifies the hypotheses of \Cref{prop:primitive-verification}.  The mean and median conclusions then follow from \Cref{thm:bootstrap,cor:fixedB,cor:functionals}.
\end{proof}

\begin{proof}[Proof of \Cref{cor:median}]
At $x_\star$, the target innovation is
\[
 \xi=\Psi(x_\star;Y)=-\frac{\Log_{x_\star}(Y)}{d(x_\star,Y)}.
\]
The no-atom condition makes this well defined almost surely, and the first-order condition for the unique differentiable population median gives $\Exp\xi=0$. The innovation has norm one and covariance $Q_{\mathrm{med}}$.  Hence the innovation covariance and Lindeberg conditions hold without a moment assumption beyond measurability.

It remains to verify the local $L^2$ modulus despite samplewise nonsmoothness. Bound the equality convention for $\Psi(x;y)$ by $M\ge1$, enlarging the bound if necessary.  On a separable compact local ball, define the measurable supremum envelope
\[
 G_t(y)=\sup_{d(x,x_\star)\le t}
 \norm{\mathcal P_{x\to x_\star}\Psi(x;y)-\Psi(x_\star;y)}^2;
\]
the supremum may equivalently be taken over a countable dense subset, with equality points included separately.  For every fixed $y\ne x_\star$, eventually $t<d(y,x_\star)/2$; smoothness away from the diagonal and the shrinking compact ball then give $G_t(y)\to0$.  Uniformly, $G_t(y)\le(M+1)^2$.  Since $\Prob(Y=x_\star)=0$, dominated convergence yields $\Exp G_t(Y)\to0$, and hence
\[
 \sup_{d(x,x_\star)\le t}
 \Exp\norm{\mathcal P_{x\to x_\star}\Psi(x;Y)-\Psi(x_\star;Y)}^2
 \le \Exp G_t(Y)\longrightarrow0.
\]
Centering the increment changes the bound by at most a constant factor, so \eqref{eq:chi-L2} holds for both original and weighted predictable arguments.  Non-atomicity on the localized neighborhood excludes their equality events.  The population $C^{1,\rho}$ assumption gives \eqref{eq:r0-holder}, and bounded score norm gives the oracle second-moment condition.  The stated inequality $\alpha(1+\rho)>1$, together with the state-rate and geometry assumptions, now allows \Cref{prop:primitive-verification} to be applied.  \Cref{thm:joint} yields \eqref{eq:median-limit}, and the bootstrap conclusions follow from \Cref{thm:bootstrap,cor:fixedB,cor:functionals}.
\end{proof}

\begingroup\begin{proof}[Proof of \Cref{prop:median-model}]
The density is smooth, bounded, compactly supported, and positive near $\mathrm{A}$. The affine-invariant symmetry $s_{\mathrm{A}}(\mathrm{y})=\mathrm{A}\mathrm{y}^{-1}\mathrm{A}$ is an isometry preserving this density and Riemannian volume, and reverses $\Log_{\mathrm{A}}(\mathrm{y})$. Thus the population normalized score vanishes at $\mathrm{A}$.

Write $r_y(x)=d(x,y)$. On compact state and observation sets, smooth normal coordinates near the diagonal and smoothness away from it give
\[
 \norm{\nabla^2r_y(x)}\le C(1+r_y(x)^{-1}),\qquad
 \norm{\nabla_x\nabla^2r_y(x)}\le C(1+r_y(x)^{-2}).
\]
These follow by differentiating the local distance expression: its first, second, and third derivatives have orders $1$, $r^{-1}$, and $r^{-2}$, respectively; bounded metric derivatives on a compact enlargement control the remaining terms. Uniform volume comparison on that enlargement bounds the integral of $r^{-2}$ over a ball of radius $\varepsilon$ by $C\int_0^\varepsilon r^{d-3}\,dr$. Since $d\ge3$ and the density is bounded, this tends to zero uniformly in the state. Differentiation of the expected score is justified by first excluding such a ball and then taking its radius to zero. Applying the fundamental theorem of calculus to the Hessian along state geodesics and Fubini's theorem gives a locally Lipschitz population derivative. Observation points lying on a fixed state geodesic form a volume-null set and do not affect that argument.

Hadamard Hessian comparison gives, for $y\ne x$,
\[
 \Hess r_y(x)[v,v]\ge
 \frac{\norm v_x^2-\ip{v}{e_{x,y}}_x^2}{d(x,y)},\qquad
 e_{x,y}=\Log_x(y)/d(x,y).
\]
The density is positive on an open set, so the expectation is strictly positive for every nonzero $v$. Continuity and compactness of the unit tangent bundle over $K$ give a uniform positive lower bound there. The expected distance is strictly geodesically convex; its stationary point $\mathrm{A}$ is the unique minimizer. Integrating its Hessian along the geodesic from $\mathrm{A}$ to any $x\in K$ yields \eqref{eq:strong-monotonicity}. The same open support makes the target normalized-score covariance positive definite.

The sample score is bounded by one, has zero approximation bias, and has a local $L^2$ modulus by the dominated-convergence argument in the proof of \Cref{cor:median}. The population remainder has $\rho=1$. The projection safeguard satisfies every requirement of \Cref{prop:state-rate}, which therefore proves the state rate and almost sure convergence. Exponential averaging satisfies \Cref{lem:averaging-basin} with $\kappa=C=1$. On a Hadamard manifold all logarithms exist, compactness gives finite coordinates, and both states and full averages eventually enter every comparison neighborhood of $\mathrm{A}$. Exponential-map geometry and \Cref{prop:primitive-verification} now establish the claimed bootstrap conclusions.
\end{proof}\endgroup

\begin{proof}[Proof of \Cref{cor:grassmann}]
Work in the standard horizontal tangent representation at $\mathrm{U}_\star$.  A tangent perturbation has the form $\mathrm{U}_\perp \mathrm{K}$.  The population Riemannian gradient of \eqref{eq:pca-objective} is
$G(\mathrm{U})=-(\mathrm{I}-\mathrm{U}\mathrm{U}^{\trans})\mathrm{\Sigma} \mathrm{U}$.
Differentiating at $\mathrm{U}_\star$ in direction $\mathrm{U}_\perp \mathrm{K}$ and using
$\mathrm{\Sigma} \mathrm{U}_\star=\mathrm{U}_\star\mathrm{\Lambda}_1$ and
$\mathrm{\Sigma} \mathrm{U}_\perp=\mathrm{U}_\perp\mathrm{\Lambda}_2$ gives
\begin{equation*}
  DG(\mathrm{U}_\star)[\mathrm{U}_\perp \mathrm{K}]
  =\mathrm{U}_\perp(\mathrm{K}\mathrm{\Lambda}_1-\mathrm{\Lambda}_2\mathrm{K})
  =\mathrm{U}_\perp\mathcal H(\mathrm{K}).
\end{equation*}
Every tangent eigenvalue is $\lambda_j-\lambda_{r+a}>0$, so $\mathcal H$ is positive definite.

At the target,
\[
  \Psi(\mathrm{U}_\star;Y)
  =-(\mathrm{I}-\mathrm{U}_\star \mathrm{U}_\star^{\trans})YY^{\trans}\mathrm{U}_\star
  =-\mathrm{U}_\perp Y_2Y_1^{\trans}.
\]
The vectorized coordinate innovation is therefore $\xi=-\vecop(Y_2Y_1^{\trans})$. It is centered because
$\mathrm{U}_\perp^{\trans}\mathrm{\Sigma} \mathrm{U}_\star=0$, and its covariance is $\mathrm{Q}_{\mathrm{PCA}}$ in \eqref{eq:pca-Q}.
The score obeys $\norm{\Psi(\mathrm{U};Y)}\le\norm Y^2$, and its local Lipschitz envelope is bounded by $C\norm Y^2$.  Thus $\Exp\norm Y^4<\infty$ gives both the innovation second moment and the local $L^2$ modulus.  The population field is smooth on the compact Grassmann manifold and has a quadratic remainder.  The state-rate and second-order retraction assumptions complete the hypotheses of \Cref{prop:primitive-verification}.  Equation \eqref{eq:pca-limit} follows from \Cref{thm:joint}; the bootstrap conclusions follow from \Cref{thm:bootstrap,cor:fixedB,cor:functionals}.

If $Y$ is Gaussian in the eigenbasis, $Y_1$ and $Y_2$ are independent with diagonal covariance matrices $\mathrm{\Lambda}_1$ and $\mathrm{\Lambda}_2$.  Hence
\[
  \Cov(Y_{2,a}Y_{1,j},Y_{2,b}Y_{1,\ell})
  =\lambda_{r+a}\lambda_j\ind(a=b)\ind(j=\ell).
\]
The inverse of \eqref{eq:pca-H} divides coordinate $(a,j)$ by
$\lambda_j-\lambda_{r+a}$, proving \eqref{eq:pca-gaussian-variance}.
For $P(\mathrm{U})=\mathrm{U}\mathrm{U}^{\trans}$,
$dP[\mathrm{U}_\perp \mathrm{K}]=\mathrm{U}_\perp \mathrm{K}\mathrm{U}_\star^{\trans}+\mathrm{U}_\star \mathrm{K}^{\trans}\mathrm{U}_\perp^{\trans}$.  Taking the trace against symmetric $\mathrm{A}$ gives \eqref{eq:projection-delta}.
\end{proof}

\subsubsection{Proof of the Approximation-Bias Proposition}\label{app:bias}

\begin{proof}[Proof of \Cref{prop:bias}]
Whenever the coordinate Abel rearrangement is valid, its nonnegligible-bias term is
\begin{equation}\label{eq:bias-base-term}
  -H^{-1}\frac1{\sqrt n}\sum_{k=1}^nb_k.
\end{equation}
For $0<\gamma<1$,
\begin{equation*}
  \sum_{k=1}^nk^{-\gamma}
  =\frac{n^{1-\gamma}}{1-\gamma}\{1+o(1)\},
\end{equation*}
while $n^{-1/2}\sum_{k=1}^nk^{-1/2}\to2$.  If $\gamma=1$, the sum is $O(\log n)$, and if $\gamma>1$, it is bounded.  Therefore the leading part of \eqref{eq:bias-base-term} tends to zero when $\gamma>1/2$, tends to $-2h_0^rH^{-1}B_0$ when $\gamma=1/2$, and equals \eqref{eq:subcritical-shift} when $0<\gamma<1/2$.  The remainder $\widetilde b_k$ is negligible by assumption.  In the first two regimes the other abstract state and remainder conditions were assumed, so this Abel term combines with \Cref{lem:base-clt,lem:average-expansion,lem:random-base} as in the main theorem; at the boundary this gives \eqref{eq:critical-shift}.  In the subcritical regime the calculation is only conditional on a separate bias-aware justification of the Abel identity and does not invoke those two geometric lemmas.

The bootstrap difference contains the centered bias term
\begin{equation}\label{eq:bias-weight-term}
  -H^{-1}\frac1{\sqrt n}\sum_{k=1}^n(W_k-1)b_k.
\end{equation}
Its conditional second moment is bounded by a constant times
$n^{-1}\sum_{k=1}^n\norm{b_k}^2$.  For the leading term in \eqref{eq:bias-polynomial}, this average is
$O(n^{-2\gamma})$ if $0<\gamma<1/2$,
$O((\log n)/n)$ if $\gamma=1/2$, and
$O(n^{-1})$ if $\gamma>1/2$.  The squared-remainder condition and Cauchy--Schwarz control the remaining terms.  Hence \eqref{eq:bias-weight-term} is $o_{\Prob^\star}(1)$ for every $\gamma>0$.  For $\gamma\ge1/2$, the remaining terms in the random-base difference are those in the proof of \Cref{thm:bootstrap}; \Cref{lem:multiplier-clt} therefore gives \eqref{eq:bootstrap-BL-new}, including the centered conditional law at the boundary.  This proves the fixed boundary statement under the abstract conditions and the stated coordinate-level diagnostic below the boundary; it does not turn the latter into a manifold-bootstrap theorem.
\end{proof}

\end{document}